\documentclass[review]{elsarticle}

\usepackage{lineno,hyperref}
\modulolinenumbers[5]

\usepackage{geometry}
\journal{Journal of Information Sciences}
\usepackage{array}
\usepackage{booktabs}
\usepackage{times}
\usepackage{soul}
\usepackage{url}
\usepackage[utf8]{inputenc}
\usepackage{bbding}
\usepackage{graphicx}
\usepackage{amsmath}
\usepackage{graphbox}
\usepackage{booktabs}
\usepackage{algorithm}
\usepackage{algorithmic}
\usepackage{lineno}
\usepackage{pifont}
\usepackage{framed}
\usepackage{graphicx}
\usepackage{subfigure}
\usepackage{multirow}
\usepackage{amsmath}
\usepackage{amsfonts}
\usepackage{color}

\begin{document}
\begin{frontmatter}

\title{Excitement Surfeited Turns to Errors: Deep Learning Testing Framework Based on Excitable Neurons}

\author[mysecondaryaddress]{Haibo Jin}
\ead{2112003035@zjut.edu.cn}

\author[mysecondaryaddress]{Ruoxi Chen}
\ead{2112003149@zjut.edu.cn}

\author[mymainaddress,mysecondaryaddress]{Haibin Zheng}
\ead{haibinzheng320@gmail.com}

\author[mymainaddress,mysecondaryaddress]{Jinyin Chen\corref{mycorrespondingauthor}}
\cortext[mycorrespondingauthor]{Corresponding author}
\ead{chenjinyin@zjut.edu.cn}


\author[myfourthaddress]{Yao Cheng}
\ead{chengyao101@huawei.com}

\author[yuyueaddress]{Yue Yu}
\ead{yuyue@nudt.edu.cn}

\author[liuxianglongaddress]{Xianglong Liu}
\ead{xlliu@nlsde.buaa.edu.cn}

\address[mysecondaryaddress]{College of Information Engineering, Zhejiang University of Technology, Hangzhou, China}
\address[mymainaddress]{Institute of Cyberspace Security, Zhejiang University of Technology, Hangzhou, China}


\address[myfourthaddress]{Huawei International, Singapore}

\address[yuyueaddress]{National Laboratory for Parallel and Distributed Processing College of Computer, National University of Defense Technology, Changsha, China}

\address[liuxianglongaddress]{State Key Laboratory of Software Development
Environment, Beihang University, Beijing, China}

\begin{abstract}

Despite impressive capabilities and outstanding performance, deep neural networks (DNNs) have captured increasing public concern about their security problems, due to their frequently occurred erroneous behaviors. Therefore, it is necessary to conduct a systematical testing for DNNs before they are deployed to real-world applications. 
Existing testing methods have provided fine-grained metrics based on neuron coverage and proposed various approaches to improve such metrics. However, it has been gradually realized that a higher neuron coverage does \textit{not} necessarily represent better capabilities in identifying defects that lead to errors. Besides, coverage-guided methods cannot hunt errors due to faulty training procedure. So the robustness improvement of DNNs via retraining by these testing examples are unsatisfactory. 
To address this challenge, we introduce the concept of excitable neurons based on Shapley value and design a novel white-box testing framework for DNNs, namely \emph{DeepSensor}.
It is motivated by our observation that neurons with larger responsibility towards model loss changes due to small perturbations are more likely related to incorrect corner cases due to potential defects.  
By maximizing the number of excitable neurons concerning various wrong behaviors of models, DeepSensor can generate testing examples that effectively trigger more errors due to adversarial inputs, polluted data and incomplete training.
Extensive experiments implemented on both image classification models and speaker recognition models have demonstrated the superiority of DeepSensor.
Compared with the state-of-the-art testing approaches, DeepSensor can find more test errors due to adversarial inputs ($\sim\times1.2$), polluted data ($\sim\times5$) and incompletely-trained DNNs ($\sim\times1.3$). Additionally, it can help DNNs build larger $l_2$-norm robustness bound ($\sim\times3$) via retraining according to CLEVER's certification.
We further provide interpretable proofs for effectiveness of DeepSensor via excitable neurons identification and visualization of t-SNE. The open source code of DeepSensor can be downloaded at~\url{https://github.com/Allen-piexl/DeepSensor/}.

\end{abstract}


\end{frontmatter}


\section{Introduction}
Deep neural networks (DNNs) have achieved tremendous progress over the last few decades and enjoyed increasing popularity in a variety of applications such as computer vision~\cite{zheng2019denoising, liang2018detecting,AgarwalSVR21,ma2018efficient} and audio recognition~\cite{QianDHCJL21,LuLHL20}, etc. Despite their impressive capabilities and outstanding performance, DNNs' security and robustness have raised massive public concern. DNNs often expose unexpected erroneous behaviors, which may finally lead to disastrous consequences especially in safety-critical applications such as autonomous driving~\cite{lambert2016understanding}. Therefore, it is important to systematically test DNNs before deploying them to safety and security critical domains. 

Extensive efforts have been made to test vulnerabilities of DNNs. 
Based on the observation of external or internal behaviors, testing methods for DNNs can be categorized into black-box and white-box testing techniques. 
The former generate testing examples without observing the internal behaviors of models~\cite{wicker2018feature, ma2018deepmutation}. 
White-box DNN testing, the mainstream of existing testing methods, is inspired by the success of code coverage criteria in traditional software testing. It was first proposed in DeepXplore~\cite{pei2017deepxplore}. Such testing methods are mainly coverage-criterion guided, which focus on improving neuron coverage in testing example generation via well-established mutation or differential process. They assumed that a better-tested DNN with higher coverage, is more likely to be robust. Along this direction, DLFuzz~\cite{guo2018dlfuzz}, DeepHunter~\cite{xie2019deephunter} and ADAPT~\cite{lee2020effective} are developed, further achieving higher coverage and deeper the exploratory degree.

However, higher neuron coverage does \textit{not} indicate better effectiveness and defect-finding capability of testing approaches~\cite{li2019structural,harel2020neuron}, due to significant difference between DNNs and software code. Besides, the limited connection between coverage and model robustness has been demonstrated~\cite{dong2019there}. So existing methods pursuing high coverage do \textit{not} necessarily find more defects due to adversarial inputs from diverse categories, nor better improve the model robustness via retraining. 
Moreover, existing coverage-guided approaches are based on activation boundary calculated from training data of well-trained models, they may fail to find defects due to faulty training procedure, i.e., errors from training with polluted data and incomplete training procedure (underfitting or overfitting). The overlook of these errors will also lead to catastrophe in the real world. 

In a word, existing white-box testing methods are still limited in terms of both \emph{(i)} testing performance-guided by coverage from well-trained models, they fail to hunt diverse test errors by adversarial inputs and those due to faulty training procedure, leading to unsatisfactory defect-finding capability; \emph{(ii)} robustness improvement, due to limited relationship between coverage and robustness, they may not improve the robustness of DNNs after retraining. 

Due to the decision results of DNNs are determined by the nonlinear combination of each neuron’s activation state, we reconsider the potential defects of DNNs from the relationship between model prediction changes and neuron behaviors. We assume that erroneous model predictions are caused by some neurons, i.e., neurons that contribute more to model prediction changes are more likely to trigger erroneous corner cases.

Inspired by cooperative game theory~\cite{shapley1953value} that allocates the benefits to various participants, we adopt Shapley value~\cite{ChenSWJ19, LiKLCZSX21} to quantitatively measure neuron's contribution towards model prediction changes due to input perturbations. Different from Guan et al.~\cite{guan2022few} that use backdoor attack success rate for calculating Shapley value in backdoored models, we adopt changes of model loss due to manually-added random noise for measurement. In this process, neurons responsible for erroneous model behaviors can be targeted. 
Based on that, we propose the concept of \emph{excitable neuron}, which is determined by Shapley value of neurons. More specifically, neurons with Shapley value larger than properly-set thresholds (e.g., 0.5 in this experiment) are considered as excitable neurons.


\begin{figure}[t]
\centering
    \includegraphics[width=0.7\linewidth]{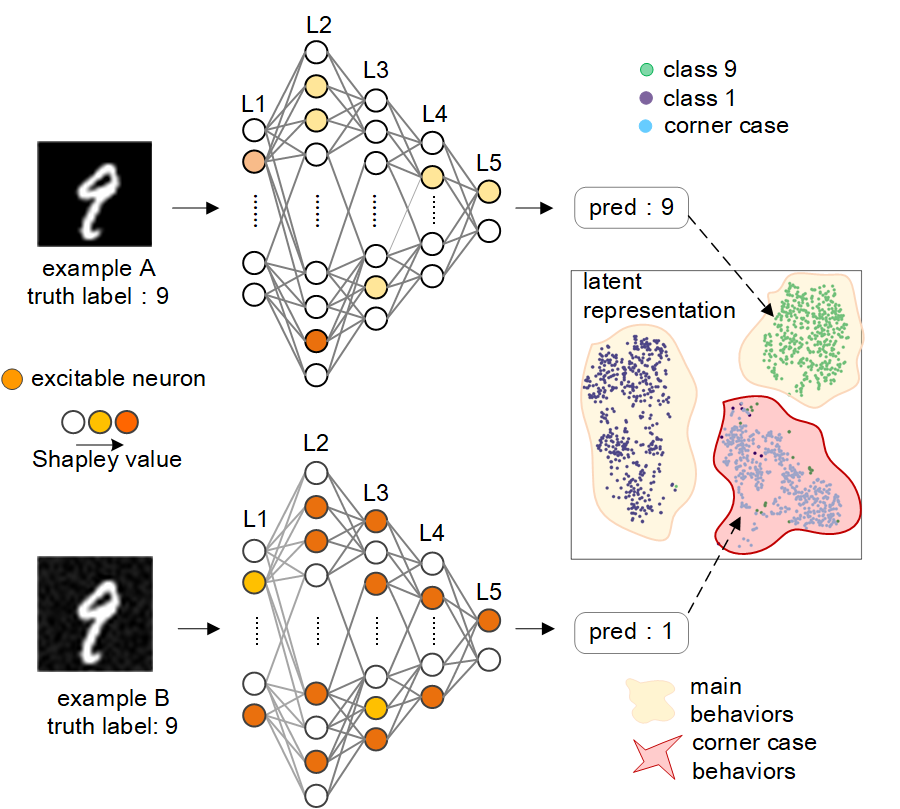}
\caption{The relationship between excitable neurons and corner case behaviors on LeNet-5 of MNIST. }
\label{observation}
\end{figure}



We conduct an experiment to verify our assumption. 
It uses the same well-trained LeNet-5~\cite{lecun2015lenet} of MNIST~\cite{lecun1998gradient} on 500 pairs of examples, i.e., a benign example from class ``9'' and one of its corner case versions. 
Fig.~\ref{observation} shows one of such example pairs.
We observe that the number of excitable neurons that are activated by the corner case example B is much more than that of the benign example A.
For better understanding, we visualize the latent representations of examples from class ``1'', class ``9'' and the corner cases using t-SNE~\cite{van2008visualizing}.
We can see that they form three different clusters, and examples in the corner case cluster generally activate more excitable neurons than that from the other clusters. 

From this experiment, we gain the insight that an input example that activates more excitable neurons in the model may expose more unexpected corner case behaviors of that model. So more potential flaws in the model can be explored if these neurons are directly targeted.

Therefore, in this paper, we leverage excitable neurons and design a novel white-box testing framework DeepSensor to hunt diverse potential defects inside the model. First, we determine excitable neurons by adding small perturbations to the benign seed. By maximizing the number of excitable neurons, perturbations are optimized during fuzzing via particle swarm optimization (PSO)~\cite{eberhart1995new}, which finds the approximate optimal solution through the collaboration and information sharing between individuals in the group. Since model loss changes are closely-related to erroneous predictions and improper training, so excitable neurons can help induce potential flaws occur due to adversarial inputs and faulty training procedure. Besides, calculating from prediction changes due to small perturbations, excitable neurons can well reflect model robustness. Consequently, through retraining by DeepSensor's testing examples, robustness can be further improved.

 

Our contributions are summarized as follows.
\begin{itemize}
\item We introduce Shapley value for measuring neuron's contribution to model prediction changes due to small perturbations. Then we identify neurons with larger Shapley value as excitable neurons. We observe that excitable neurons are closely related to the model's incorrect corner case behaviors due to various potential defects.
\item We use excitable neurons to guide DNN's testing and develop a novel white-box testing framework DeepSensor. By maximizing the number of excitable neurons via PSO, DeepSensor can generate testing examples that trigger more unexpected corner case behaviors due to different causes.
\item Compared to the SOTA testing approaches, DeepSensor can find more test errors due to adversarial inputs ($\sim\times1.2$) and polluted data ($\sim\times5$). Besides, it can help build more robust DNNs with larger $l_2$-norm robustness bound ($\sim\times3$) of CLEVER via retraining. We also demonstrate the performance of DeepSensor on speaker recognition models.
\item Based on our insights, we further provide interpretable proofs for effective testing via excitable neurons. Visualizations of DeepSensor also confirms its superiority. 
\end{itemize}

The rest of the paper is organized as follows. Related works are introduced in Section \ref{related}, while preliminaries and proposed method are detailed in Section \ref{pre} and \ref{method}. Experiments settings are expressed in Section \ref{sec_setting} while results and analysis are showed in Section \ref{exp}. Finally, we discuss limitations and conclude our work.

\section{Related Work\label{related}}
This section, we will briefly introduce testing methods for DNNs, cooperative game theory in Deep Learning (DL) systems and optimization algorithms. 
\subsection{Testing Methods for DNNs}
Testing methods are mainly designed to find vulnerabilities of DNNs before deployment. Generally speaking, according to the observation of external or internal behaviors, they can be divided into black-box and white-box methods.
\subsubsection{Black-box Methods}
Black-box testing methods achieve testing by observing external behaviors of models. 

Wicker et al.~\cite{wicker2018feature} used scale invariant feature transform to extract key points and generate testing examples, whose effectiveness is demonstrated on adversarial attacks. Instead of observing model internal behaviors, DeepMutation~\cite{ma2018deepmutation} mutates DNNs (i.e., injecting faults either from the source-level or model-level) to evaluate the quality of test data, useful for test data prioritization in respect of robustness. Wang et al.~\cite{wang2021robot} proposed a light-weighted robustness metric based on first-order loss, and used it for generating testing examples for improving model robustness. Wang et al.~\cite{wang2022bet} proposed a black-box efficient testing method (BET), which contains a tunable objective function to explore defects in different decision boundaries of DNNs.

\subsubsection{White-box Methods}
Different from black-box methods, white-box ones reach the objective from the  perspective of model inside. They are mainly based on coverage, which can be further divided into neuron coverage and multi-granularity neuron coverage.

\textbf{Neuron coverage based.} Pei et al.~\cite{pei2017deepxplore} designed the first white-box testing framework DeepXplore and introduced neuron coverage to systematically finding inputs that can trigger corner case behaviours between multiple models. It randomly chooses neurons and generates testing examples via gradient descent. However, it shows limitation on measuring all neuron behaviors with one single metric. Based on it, Guo et al.~\cite{guo2018dlfuzz} proposed DLFuzz based on differential fuzzing, which mutates the input to maximize neuron coverage and the difference between original inputs and mutations. Experiments show that it reaches higher neuron coverage with less time, when compared with DeepXplore. Tian et al.~\cite{tian2018deeptest} studied a basic set of image transformations based on neuron coverage and proposed DeepTest. It shows powerful malicious behavior detection efficiency on autopilots. Besides, Odena et al.~\cite{odena2019tensorfuzz} developed coverage-guided fuzzing method TensorFuzz, to find errors that only occur on rare inputs. The random mutation of the input is guided by the coverage criteria to achieve the goal of the specified constraints. Yu et al.~\cite{yu2022white} proposed Test4Deep, an effective white-box testing DNN approach based on neuron coverage. Zhang et al.~\cite{zhang2021cagfuzz} trained an adversarial example generator to craft testing examples. They proposed CAGFuzz for detecting hidden errors in the target DNN model under the guidance of neuron coverage.

\textbf{Multi-granularity neuron coverage based.} Ma et al.~\cite{ma2018deepgauge} put forward a set of multi-granularity test standards for DNNs at a more fine-grained level, including neuron coverage (NC), strong neuron activation coverage (SNAC), \emph{k}-multisection neuron coverage (TMNC), top-\emph{k} neuron patterns (TKNP), top-\emph{k} neuron coverage (TKNC) and neuron boundary coverage (NBC). Following them, Xie et al.~\cite{xie2019deephunter} proposed DeepHunter, which uses multi-granularity criterion feedback to guide the generation of testing example using mutation-based optimization. As a result, it realizes further detection of potential corner cases. Based on NC and TKNC, Lee et al.~\cite{lee2020effective} proposed ADAPT that can adaptively determine neuron-selection strategies during testing. Xie et al.~\cite{xie2022npc} proposed neuron path coverage (NPC) based on the decision logic of DNNs to test errors. They believed that the higher the path coverage, the more diverse decision logic the DNN is expected to be explored.


\textbf{Limitations of coverage-based metrics.} The fore-mentioned research assumes that increasing coverage-based metrics can improve the quality of test suites. However, some works~\cite{harel2020neuron, li2019structural, dong2019there, yan2020correlations, pavlitskaya2022neuron} have found that using coverage-based metrics to assess DNN behaviors may be misleading. High coverage cannot indicate the effectiveness of testing approaches. More specifically, Harel et al.~\cite{harel2020neuron} invoked skepticism that increasing neuron coverage may not be a meaningful objective for generating testing examples. They also verified that taking neuron coverage into consideration actually led to fewer test inputs found. Yan et al.~\cite{yan2020correlations} found that coverage criteria are not correlated with model robustness. Pavlitskaya et al.~\cite{pavlitskaya2022neuron} found no evidence that the investigated coverage-based metrics can be advantageously used to improve robustness.


\subsection{Cooperative Game Theory in DL Systems\label{Cooperative}}
Cooperative game theory~\cite{shapley1953value} studies how to allocate worth to cooperative participants when they reach cooperation. Recently, many researchers tried to use cooperative game theory for better understanding how deep learning systems work. Shapley value, derived from cooperative game theory, is a well-known distribution method based on contributions.

Zhang et al.~\cite{zhang2020interpretingandboosting} explained the utility of dropout using the interaction defined in cooperative game theory, and revealed the close relationship between the strength of interactions and the over-fitting of DNNs. Ghorbani et al.~\cite{ghorbani2020neuron} used Shapley value to quantify the response of individual neurons to the model performance, rather than activation patterns that are used by others. Ren et al.~\cite{ren2021unified} discovered that adversarial attacks mainly affect high-order interactions between input variables by cooperative game theory. In NLP models, Zhang et al.~\cite{zhang2021building} used interactions based on Shapley values of words for building up a tree to explain the hierarchical interactions between words. Li et al.~\cite{li2021instance} proposed COncept-based NEighbor Shapley approach (CONE-SHAP), which evaluates the importance of each concept by considering its physical and semantic neighbors. They explained the model knowledge with both instance-wise and class-wise explanations. Lu et al.~\cite{lu2022interpretable} utilized the Shapley interaction index as the strategy to measure the information gain of image pixels. Based on it, they can detect the suspicious tampered region. 
In the security domain, Guan et al.~\cite{guan2022few} used Shapley value to locate neurons that are the most responsible for backdoor behaviors. Based on it, they proposed Shapley Pruning framework to detect and mitigate backdoor attacks from poisoned models. Different from them that use backdoor attack success rate to target poisoned neurons, we focus on model prediction changes for searching excitable neurons.

\subsection{Optimization Algorithms\label{opt}}
In this part, we introduce several commonly-used optimization algorithms.

Inspired by the foraging behavior of some ant species, Dorigo et al.~\cite{DorigoBS06} proposed ant colony optimization to stimulate the process of ant foraging. It is easy to combine with other methods, but takes high computation cost. 
Genetic algorithm~\cite{BurkeV97} is motivated by the evolutionary laws in nature, which searches for best solution via three genetic operators, including selection, crossover and mutation. Its robustness and scalability has been demonstrated in many applications, with problems of low efficiency and complex parameters left to be solved. 
PSO is an evolutionary computation approach developed by Kennedy et al.~\cite{eberhart1995new}. In PSO, each particle can record the historical optimal position, and the group can be optimized through the information sharing between the particles. It converges fast with fewer parameters, but easy to stuck at local optima. Considering low time cost and stable performance, PSO is adopted for optimization during testing example generation in DeepSensor. 




\section{Preliminaries\label{pre}}

\subsection{Problem Definition}
A DNN with potential flaws will expose erroneous behaviors when faced with uncertain inputs. Here we list three goals we attempt to achieve through testing. (i) To find the correlation between neurons and wrong behaviors, based on observations of changes of the model loss.
Then we can interpret the reason for corner case behaviors. (ii) On the basis of our insights, we generate testing examples that can trigger more erroneous behaviors. Both diversity and efficiency are taken into consideration in this process. (iii) We implement testing on various datasets for demonstration and further improve model robustness via retraining.

\subsection{Model Loss And Robustness}
A DNN with parameter $\theta$ can be denoted as $f_{\theta}(x): \mathcal{X}\rightarrow \mathcal{Y}$, where $x \in X \subset \mathbb{R}^N$ represents the input and $y \in Y$ denotes predicted labels. Given a dataset $D={(x_i,y_i)}^m_{i=1}$ with $m$ examples, during traning, the loss function of this DNN is calculated as:
\begin{equation}
    \mathcal{L}(x,y,\theta)= \frac{1}{m} \sum^m_{i=1} \mathcal{J}(f_{\theta}(x_i),y_i) = -\frac{1}{m} \sum^m_{i=1} y_i~\text{log}(f_{\theta}(x_i))
\end{equation}
where $\mathcal{J}$ calculates the cross-entropy between the model output $f_{\theta}(x_i)$ and the ground truth label $y_i$. $\text{log}(\cdot)$ is the logarithmic function. Model parameter $\theta$ will be updated until the training process terminates. In the following, we omit $\theta$ for model $f(x)$ after training. 

Given an input $x$ and a small perturbation $\Delta x$, the robustness of $f(x)$ can be theoretically formalized as:
\begin{equation}
    \forall~ x\in X, ||\Delta x||_p \leq \epsilon \Leftrightarrow ||f(x+\Delta x)-f(x)|| \leq \delta
\end{equation}
where ($\epsilon, \delta$) denotes small numbers. Intuitively, models with better robustness have smaller $\delta$.

When $\Delta x$ is added on $x$, we formalize the change of model predictions, i.e., the change of model loss as: 
\begin{equation}
\label{loss_robustness}
   \begin{aligned}
    \Delta \mathcal{L}(x,\Delta x)=&\mathcal{L}(x+\Delta x,y)-\mathcal{L}(x,y)\\
    =& -\frac{1}{m} \sum^m_{i=1}y_i\text{log}f(x_i)+\frac{1}{m} \sum^m_{i=1}y_i\text{log}f(x_i+\Delta x_i)\\
    =& \frac{1}{m} \sum^m_{i=1}y_i ~ \text{log}\frac{f(x_i+\Delta x_i)}{f(x_i)}
   \end{aligned}
\end{equation}
From the equation, we bridge the gap between loss changes and robustness, i.e., models that merely change predictions are robust towards small perturbations. 

Numerous practical certifications of robustness have been proposed, e.g., global Lipschitz constants~\cite{szegedy2013intriguing}, $\rm AI^2$~\cite{gehr2018ai2} and CLEVER~\cite{weng2018evaluating}. Among them, CLEVER is the first attack-independent robustness metric that are feasible to general DNNs. So we use it to evaluate the robustness improvement in our experiment.

\subsection{Shapley Value of Neurons}
Each DNN consists of multiple layers: an input layer, an output layer, and many hidden layers. Each layer contains many neurons. The output of the layer is the nonlinear combination of each neuron’s activation state in the previous layer. During forward propagation, the combinations of each neuron finally determine DNNs' decision results. Each neuron can be treated as a participant playing different roles in model predictions. Therefore, the DNN with multiple neurons can be regarded as a n-player game, where we can use Shapley value~\cite{ChenSWJ19, LiKLCZSX21} to measure the contribution of each neuron.

Let $X=\{x_1, x_2, ...\}$ denote a set of benign examples. We add small perturbations $\Delta x$ on the benign example $x \in X$ and use $\Delta \mathcal{L}(x,\Delta x)$ to measure model prediction changes. Although complex interactions may exist between different neurons, we intend to assign responsibility of prediction changes to each neuron $n \in N$. We allocate contribution of each neuron by its marginal contribution.

\textbf{Definition 1: Marginal contribution.} Let $N=\{n_1, n_2,...\}$ be a set of neurons in the DNN and $s$ denotes an arbitrary subset of $N$. For neuron $n \in s$, the marginal contribution of it can be defined as:
\begin{equation}
\label{marginal contribution}
    m(n,s)=\varsigma (s)-\varsigma(s\setminus\{n\})
\end{equation}
where $\varsigma (\cdot)$ is the utility function, and here we use $\Delta \mathcal{L}(x,\Delta x)$ for calculating neuron marginal contribution. $s\setminus \{n\}$ denotes the removal of $n$, i.e., we set its activation value to zero.

\textbf{Definition 2: Shapley value of neurons.} Shapley value measures the contribution of several participants working in a big coalition. Shapley value $\psi(n)$ of neuron $n$ can be calculated by taking average of the marginal contribution:
\begin{equation}
\label{shapley}
    \psi(n) =\sum_{s\in N}\omega(|s|)m(n,s)
\end{equation}
where $\omega (\left |  s\right | )=\frac{(\left | s \right |-1 )!(\left | N \right |-\left | s \right |)!}{\left | N \right |!} $ represents the relative importance
of set $N$, and $\left |  \cdot \right |$ indicates the number of elements.



\textbf{Definition 3: Excitable neuron.} Given a neuron $n$ in the model and the threshold $\lambda$. If $\psi(n)>\lambda$, it is considered as excitable neuron, denoted as $n_e$. Excitable neurons with larger Shapley values towards model loss changes are more responsible for prediction changes. They can further be utilized to induce more erroneous predictions. 



\section{Methodology\label{method}}
\begin{figure}[htbp]
\centering
\setlength{\abovecaptionskip}{-1.5pt}
        \includegraphics[width=0.7\linewidth]{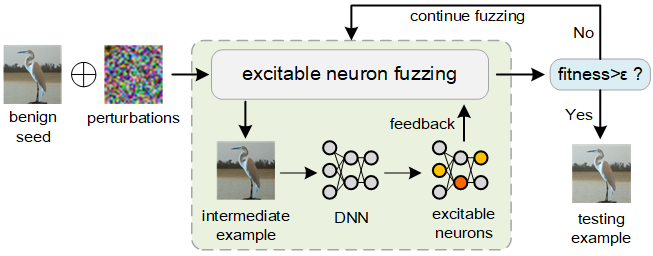}
\caption{The workflow of DeepSensor for testing DNNs.}
	\label{fig:framework}
\end{figure}

This section introduces the design and implementation of DeepSensor,
which aims to test DNNs automatically. The workflow of it is shown in Fig.~\ref{fig:framework}. Intermediate examples denote examples that are generated during fuzzing but not chosen as final testing examples. Excitable neurons are used to guide fuzzing of testing examples. In this process, PSO is used for optimization of perturbations.


More specifically, benign seeds are the input of DeepSensor. After adding small perturbations, these examples seeds compose intermediate examples, which will be fed into the DNN. Excitable neurons are identified and targeted and the number of them are used in the fitness function during the fuzzing process. The iteration terminates when intermediate examples can activate enough excitable neurons that reach the fitness function, or the maximum number of iterations is reached. We treat these examples as final testing examples that can trigger corner cases.

\subsection{Excitable Neuron Fuzzing}
The guided fuzzing is the main component of DeepSensor, which is completed by maximizing the number of excitable neurons. In this part, two key components, excitable neuron selection and the fitness function, will be discussed in detail.

\subsubsection{Excitable Neuron Selection} 
We assume that neurons with larger Shapley values are more likely to trigger misclassifications. To reach the goal of optimization, we first identify and target excitable neurons. By adding small perturbations to the given benign seed $x$, we traverse each neuron in the model to calculate its Shapley value. 


\subsubsection{Fitness Function} 
Fitness function is the main basis of the iteration, which is also the guidance of generating testing examples. In this process, we maximize the number of excitable neurons. Hence, the fitness function is designed as the following equation shows: 

\begin{equation}
fitness= \frac {\sum_{i=1}^lnum(n_{e,i})}{num(N)}
\label{fitness}
\end{equation}
where $num(N)$ denotes the total number of neurons, $num(n_{e,i})$ is the number of excitable neurons in the $i$-th layer, respectively. $l$ is the total number of layers in the model. $\sum$ means the sum function. 

The ratio of excitable neurons is calculated as the value of the fitness function, which is 0 at the beginning and is supposed to increase significantly during iterations. The threshold value of fitness function is denoted as $\epsilon$. If $\epsilon$ is reached, the fuzzing process ends.

\subsection{Testing Example Generation via PSO\label{generating}}


As mentioned in Section \ref{opt}, PSO is adopted for optimization due to low computation complexity. To start, we first need to initialize the particle swarms. For image dataset, the initial position matrix $x_p$ is set to the RGB values of pixels in each seed example $x$. Similarly, for audio dataset, initial position matrix $x_p$ is set to the feature vector of the input. The velocity matrix $v$ is set to the perturbed value $\Delta x$. Particle swarms are initialized randomly, to avoid the problem of local optimum. The population size is set to 100, which means that 100 intermediate examples (i.e., particles) are kept in each iteration. 



During iterations, the position of a single particle ${x_p}= x_p+v$ denotes the intermediate example $\hat{x}=x+\Delta x$. The fitness value of each single particle ${x_p}$ is calculated according to Eq.~\ref{fitness} and the particle with best fitness value is called $p_{best}$. Similarly, the global optimal position $g_{best}$ is the best among the population of particles we kept. In each iteration,  we compare the current value with the previous best values of particle swarms in the population, and update $p_{best}$ of each particle, as well as $g_{best}$.

Next, the speed $v$ and position ${x_p}$ of the particle swarm will be updated according to the following equation. The updating formula of velocity consists of three parts, i.e., the momentum part, the individual and the group part. The momentum part is responsible for velocity maintenance, and the individual and group part guarantee the approach towards $p_{best}$ and $g_{best}$, respectively. Non-negative inertia weight is adopted in the update process to avoid the problem of local optimum. 

\begin{equation}
    \omega^g=(\omega_{initial}-\omega_{end})(G_k-g)/G_k+\omega_{end}
\end{equation}
\begin{equation}
    \begin{aligned}
    v_i=\omega^g \times v_i+c_1\times rand(\cdot)\times(p_{best}-{x_p})\\
     +c_2\times rand(\cdot)\times(g_{best}-{x_p})
    \end{aligned}
\end{equation}
\begin{equation}
    {x_p}={x_p}+v
\end{equation}
where $\omega^g$, $\omega_{initial}$ and $\omega_{end}$ denote the current, initial and final weight vector, respectively. $g$ represents the number of current iteration and $G_k$ is the maximum iteration. $c_1$ and $c_2$ are learning factors, usually are set to 2. The former is responsible for individual learning while the latter for group learning. $rand(\cdot)$ represents the function that generates random number in the interval of [0,1].

Steps above will be repeated until the value of fitness function reaches threshold $\epsilon$ or $g$ reaches the maximum iteration $G_k$. The approximate optimal solution ${x_p}$, as well as the example $x^{*}$ is used as the final testing example. The pseudo-code of generating testing examples is presented in Algorithm 1. 
\begin{figure}[htbp]
\centering
        \includegraphics[width=0.55\linewidth]{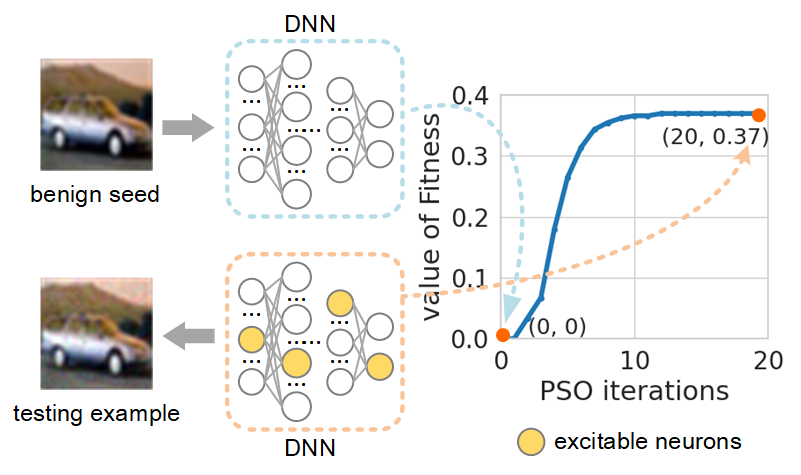}
\caption{Illustration of generating testing examples on VGG16  of CIFAR-10. The fitness function is abbreviated as ``Fitness''.}
	\label{iteration}
\end{figure}

\textbf{A running example.} We use VGG16~\cite{simonyan2014very} model on CIFAR-10~\cite{krizhevsky2009learning} dataset as a running example to illustrate our approach, as shown in Fig. \ref{iteration}. The blue curve in the right part denotes the value of the fitness function (i.e., the ratio of excitable neurons) with the increase of iterations. The maximum number of iteration $G_k$ is set to 20 and $\epsilon$ is 1.

The benign seed (classified as ``car'') is input to the DNN. At the beginning of the iteration, no excitable neurons are activated so the value of fitness function is 0. PSO changes the value of image pixels, to activate more excitable neurons in the DNN. During fuzzing, the value of the fitness function increases with the growing number of iteration. The iteration ends when $G_k$=$20$ is reached. The intermediate example is output as the testing example, which is misclassified as ``airplane''. It must be noted that here $G_k$ is reached before the threshold $\epsilon$. Actually, the value of the fitness function remains stable after 10th iteration. This indicates that in the 10th iteration, neurons that are responsible for model loss changes are all activated by the intermediate example. Finally, the generated testing example activates 37\% of the whole neurons to become excitable neurons, which can already trigger the corner case. 


\makeatletter
\makeatother
\begin{figure}[htbp]
\centering
  \renewcommand{\algorithmicrequire}{\textbf{Input:}}
  \renewcommand{\algorithmicensure}{\textbf{Output:}}
  \begin{algorithm}[H]
    \caption{Generation of testing examples}
    \begin{algorithmic}[1]
      \REQUIRE Benign seeds $X=\{x_1,x_2,…\}$. Fitness function ${Fitness(\cdot)}$ and its threshold $\epsilon$. Velocity matrix $v$. Population size $NUM$. The maximum iteration $G_k$. 
      \ENSURE Generated examples ${X}^{*}=\{x_1^*,x_2^*,…\}$.
      \STATE ${{X}^{*}=\{\emptyset\}}$
      \FOR {$x$ in $X$}
      \STATE $p_{best}$,$g_{best}$=\textbf{InitializeSwarm($x$)}
      \FOR {current iteration $g$ $\leq$ $G_k$}
      \STATE {Intermediate example $\hat{x}$=$x+v$}
      \STATE ${Fitness({\hat{x}})}$=\textbf{CalculateFitness($\hat{x}$)}\
      \FOR {$i$ =${0:NUM}$}
      \IF{${Fitness(\hat{x})}$ \textgreater $p_{best}$}
      \STATE {$p_{best}$=${Fitness(\hat{x})}$}
      \IF{${Fitness(\hat{x})}$ \textgreater $g_{best}$}
      \STATE {$g_{best}$=${Fitness(\hat{x})}$}
      \ENDIF
      \ENDIF
      \STATE{\textbf{CalculateSpeed($\hat{x}$)} }
      \STATE{\textbf{UpdatePosition($\hat{x}$)}}
      \ENDFOR
      \STATE{$g+=1$}
      \IF{$g_{best}$ \textgreater $\epsilon$}
      \STATE{\textbf{break}}
      \ENDIF
      \ENDFOR
      \STATE{$x^*=\hat{x}$, \textbf{the best solution ${x}^{*}$ is found}}
      \STATE{$X^* \leftarrow X^* \cup x^*$}
      \ENDFOR
    \end{algorithmic}
  \end{algorithm}
\end{figure}

\subsection{The Solution to Local Optimum of PSO}
It is generally believed that the global version of the particle swarm algorithm converges quickly, but easily falls into a local optimum. To avoid the problem, we randomly assign the initial location of each swarm before iterations. In this way, in the process of updating the particle position, global optimal can be reached with a higher probability. 
The effectiveness of this solution will be verified in Section \ref{pso_random}, where the impact of initial position of swarms in PSO is analyzed.


\section{Experimental Settings\label{sec_setting}}

\subsection{Datasets} 
We evaluate DeepSensor on three datasets related to image classification. Since speaker recognition systems are widely used in the real-world scenarios, so we also evaluate DeepSensor about speaker recognition dataset.

For the task of image classification, we use MNIST, CIFAR-10 and ImageNet. MNIST~\cite{lecun1998gradient} is a handwritten digit dataset, which contains 70,000 28×28 gray-scale images. 50,000 examples are used for training, 10,000 for testing. Each is marked with numbers from 0 to 9. CIFAR-10~\cite{krizhevsky2009learning} dataset consists of 60,000 32×32 RGB-color images, which are divided into ten classes equally. In our experiment, 60,000 examples are used for training and 10,000 for validation set. ImageNet~\cite{russakovsky2015imagenet} is a large set of conventional images for classification. In our experiment, 11,000 images are selected for training and 2,000 images for testing. 

On applications of speaker recognition, VCTK dataset~\cite{yamagishi2019cstr} is used. It includes speech data uttered by 109 native speakers of English with various accents. Each speaker reads out about 400 sentences, and our goal is to identify the speaker's accent. 10,000 examples from 30 classes are used for training, and 1,200 are for testing.

\subsection{DNNs} 
Various models are adopted in our experiments.
On MNIST, we adopt three LeNet family models~\cite{lecun2015lenet} (i.e., the LeNet-1, LeNet-4, LeNet-5), abbreviated as ``LN-1'', ``LN-4'' and ``LN-5''. ResNet-20 (Res-20)~\cite{he2016deep} and VGG16~\cite{simonyan2014very} are used for CIFAR-10. For ImageNet, two pre-trained models MobileNet (MNet)~\cite{howard2017mobilenets} and ResNet-50 (Res-50)~\cite{he2016deep} are used. ResNet-34~\cite{he2016deep} and Deep Speaker~\cite{li2017deep} on VCTK dataset are adopted for speaker recognition. Model configurations and classification accuracy of testing are detailed in Table~\ref{data_model}. Noted that ``acc" gained here is calculated on the validation set.

\begin{table}[t]
\centering
\caption{Well-trained models used in experiments and their configurations.}
\resizebox{0.65\linewidth}{!}{
\begin{tabular}{ccrcrc}
\toprule
\textbf{Datasets}                  & \textbf{Models}    & \textbf{\#Parameters} & \textbf{\#Layers} & \textbf{\#Neurons} & \textbf{acc}     \\ \hline
\multirow{3}{*}{MNIST}    & LeNet-1   & 7,206        & 7        & 52        & 98.40\% \\
                          & LeNet-4   & 69,362       & 8        & 148       & 98.90\% \\
                          & LeNet-5   & 107,786      & 9        & 268       & 98.90\% \\ \hline
\multirow{2}{*}{CIFAR-10} & ResNet-20 & 273,066      & 71       & 1,882     & 91.70\% \\
                          & VGG16    & 33,663,070   & 17       &  12,426    & 90.80\% \\ \hline
\multirow{2}{*}{ImageNet} & MobileNet    & 4,231,976  &  87       & 38,904    & 87.10\% \\
                          & ResNet-50    & 25,583,592  & 176       & 94,059    & 92.90\% \\\hline
\multirow{2}{*}{VCTK}     & ResNet-34     & 21,315,166  & 158 & 36,254 & 97.27\% \\
                          & Deep Speaker & 24,201,118  & 101 & 28,190 & 98.75\% \\  
                          \bottomrule
\end{tabular}

}
\label{data_model}
\end{table}

\subsection{Baselines\label{baseline_setting}} 
Three state-of-the-art (SOTA) testing methods are adopted for comparison with DeepSensor, including DeepXplore \cite{pei2017deepxplore},  DLFuzz \cite{guo2018dlfuzz} and DeepHunter \cite{xie2019deephunter}. These three baselines all generate testing examples via gradient descent. We obtained the implementation of these baselines from GitHub\footnote{https://github.com/peikexin9/deepxplore}\footnote{https://github.com/turned2670/DLFuzz}\footnote{https://bitbucket.org/xiaofeixie/deephunter/src/master/}. As for hyper-parameters, DeepXplore is conducted under the guidance of NC and ``blackout'' is used. We instantiated DLFuzz with the strategy that performs best, with $neuron\_to\_cover\_num$=$10$ and $k$=4. DeepHunter is calculated based on SNAC, based on which we can find more corner cases. Neuron boundary of SNAC is calculated by 1,000 examples. 

\subsection{Evaluation Metrics}
The metrics used for evaluating DeepSensor can be divided into two parts, one for evaluating basic configurations and the other for testing performance. The first includes classification accuracy (acc), while the latter includes attack success rate (ASR) and CLEVER. All metrics are defined as follows:

\textcircled{1} Classification accuracy: $acc=\frac{n_{true}}{n_{benign}}$, where $n_{true}$ is the number of benign examples correctly classified by the targeted model and $n_{benign}$ denotes the total number of benign examples.

\textcircled{2} Attack success rate: $ASR = \frac{n_{adv}}{n_{benign}}$, where $n_{adv}$ denotes the number of examples misclassified by the targeted model after attacks.

\textcircled{3} Robustness measurement CLEVER~\cite{weng2018evaluating}: CLEVER computes the lowest robust boundary ${L}_{p,x_0}^j$ through sampling a set of points $x^j$ around an input $x_0$ and takes the maximum value of $\left \| \Delta g(x_{0} ) \right \| _p$, which is calculated as: $L_{p,x_0}^{j} =\max_{x\in \mathbb{B}_p(x_0,R) } \left \| \Delta g(x_{0}) \right \| _p$, where $g(x_{0})=f_c(x_0)=f_j(x_0)$ and $c$, $j$ are two different classes.



\begin{table*}[t]
\centering
\large
\caption{Configurations of models under the setting of polluted data and incomplete training.}
\resizebox{1\linewidth}{!}{
\begin{tabular}{ccccccccccccc}
\toprule
                            & \multicolumn{8}{c}{\textbf{Polluted data} }    & \multicolumn{4}{c}{\textbf{Incomplete training}}  \\ \cline{2-13} 
                            &                                                     &                                                                                                     &                           &                                                                        &                                                                             & \multicolumn{3}{c}{Retrained acc}                                              &                           &                           &                                                                        &                                                                           \\ \cline{7-9}
\multirow{-3}{*}{\textbf{Models} }    & \multirow{-2}{*}{Patch size}                        & \multirow{-2}{*}{\begin{tabular}[c]{@{}c@{}}Polluted \\ class\end{tabular}}                         & \multirow{-2}{*}{\#Epoch} & \multirow{-2}{*}{\begin{tabular}[c]{@{}c@{}}Batch\\ size\end{tabular}} & \multirow{-2}{*}{\begin{tabular}[c]{@{}c@{}}\#Training\\ data\end{tabular}} & $\alpha$=10\%            & $\alpha$=20\%            & $\alpha$=30\%            & \multirow{-2}{*}{Config.} & \multirow{-2}{*}{\#Epoch} & \multirow{-2}{*}{\begin{tabular}[c]{@{}c@{}}Batch\\ size\end{tabular}} & \multirow{-2}{*}{\begin{tabular}[c]{@{}c@{}}Retrained\\ acc\end{tabular}} \\ \hline
                            & { min=1$\times$1,}              &                                                                                                     &                           &                                                                        &                                                                             &                          &                          &                          & overfitting               & 30                        & 32                                                                     & 99.91\%                                                                   \\
\multirow{-2}{*}{LeNet-5}   & { max=6$\times$6} & \multirow{-2}{*}{`1'}                                                                               & \multirow{-2}{*}{10}      & \multirow{-2}{*}{32}                                                   & \multirow{-2}{*}{60,000}                                                    & \multirow{-2}{*}{97.4\%} & \multirow{-2}{*}{98.3\%} & \multirow{-2}{*}{97.2\%} & underfitting              & 1                         & 512                                                                    & 79.32\%                                                                   \\ \hline
                            & { min=3$\times$ 3}              &                                                                                                     &                           &                                                                        &                                                                             &                          &                          &                          & overfitting               & 120                       & 64                                                                     & 98.20\%                                                                   \\
\multirow{-2}{*}{VGG16}     & { max=8$\times$8}               & \multirow{-2}{*}{`bird'}                                                                            & \multirow{-2}{*}{80}      & \multirow{-2}{*}{64}                                                   & \multirow{-2}{*}{60,000}                                                    & \multirow{-2}{*}{89.8\%} & \multirow{-2}{*}{88.9\%} & \multirow{-2}{*}{89.4\%} & underfitting              & 5                         & 64                                                                     & 76.03\%                                                                   \\ \hline
                            & { min=10$\times$10}             & { }                                                                             &                           &                                                                        &                                                                             &                          &                          &                          & overfitting               & 120                       & 64                                                                     & 98.74\%                                                                   \\
\multirow{-2}{*}{ResNet-50} & { max=15$\times$15}             & \multirow{-2}{*}{{ \begin{tabular}[c]{@{}c@{}}`Siberian\\ husky'\end{tabular}}} & \multirow{-2}{*}{80}      & \multirow{-2}{*}{64}                                                   & \multirow{-2}{*}{20,000}                                                    & \multirow{-2}{*}{91.2\%} & \multirow{-2}{*}{90.7\%} & \multirow{-2}{*}{90.4\%} & underfitting              & 30                        & 64                                                                     & 79.32\%                                                                   \\ \bottomrule
\end{tabular}
}

\label{setting}

\end{table*}

\subsection{Implementation Details\label{all_imple}}
To fairly study the performance of the baselines and DeepSensor, our experiments have the following settings, unless otherwise specified: (1) The parameters in DeepSensor are set as: $c_1$=$2$, $c_2$=$2$, $\omega_{initial}$=$0.4$, $\omega_{end}$=$0.9$, $\lambda$=$0.5$, $\epsilon$=$1$, and $G_k$=$10$. The population size is 100. (2) We set he number of iterations of all testing methods to 20. (3) Configurations of models under the setting of polluted data and incomplete training are shown in Table \ref{setting}. 

All experiments are conducted on a server under Ubuntu 20.04 operating system with Intel Xeon Gold 5218R CPU running at 2.10GHz, 64 GB DDR4 memory, 4TB HDD and one GeForce RTX 3090 GPU card.
\section{Evaluation and Analysis \label{exp}}
We evaluate the performance of DeepSensor with baselines following five research questions (RQs):  
\begin{itemize}
    \item RQ1: Compared with the SOTA baselines, can DeepSensor find more adversarial inputs that belong to more categories? And can it expose test errors due to faulty training procedure? 
    \item RQ2: Does DeepSensor achieve higher model robustness, when compared with baselines?
    \item RQ3: How to interpret the effectiveness of DeepSensor via excitable neurons and visualizations? 
    \item RQ4: How applicable is DeepSensor on speaker recognition models? 
    \item RQ5: How different parameters and strategies impact DeepSensor?
\end{itemize}
    
\subsection{RQ1: Effectiveness}
When reporting the results, we focus on the following aspects: finding adversarial inputs, finding test errors due to faulty training procedure on image classification models. 

\subsubsection{Finding Adversarial Inputs}
The experimental results on image classification models are shown in Table~\ref{adv input}. Both quality and diversity are taken into consideration in this process.

\textbf{Implementation Details.} (1) We run DeepSensor and baselines using 1,000 benign seeds for 20 iterations. These benign seeds are all correctly classified. (2) In this part, we take intermediate examples during fuzzing into consideration. (3) For quality measurement, we count the total number of test errors found (\#Test errors) by testing examples. For diversity measurement, the total number of error categories found by different methods (\#Error categories), and the number of error categories found per benign seed (\#Average categories) are calculated. 

\begin{table}[t]
\centering
\caption{Effectiveness in finding adversarial inputs.}
\resizebox{0.8\linewidth}{!}{
\begin{tabular}{cccccc}
\toprule
\textbf{Datasets}                   & \textbf{Models }                     &  \begin{tabular}[c]{@{}c@{}} \textbf{Methods}  \end{tabular} & { \begin{tabular}[c]{@{}c@{}}  \textbf{\#Test} \textbf{errors} \end{tabular}} & { \begin{tabular}[c]{@{}c@{}}\textbf{\#Error } \textbf{categories}\end{tabular}} & \begin{tabular}[c]{@{}c@{}}\textbf{\#Average } \textbf{categories} \end{tabular} \\ \hline                           &                            & DeepXplore                                                    & 17,528                                                                        & 2,667                                                                               & 2.667                                                          \\
                           &                             & DLFuzz                                                        & 17,490                                                                         & 1,682                                                                               & 1.682                                                          \\
                           &                             & DeepHunter                                                    & 18,124                                                                         & 2,682                                                                               & 2.682                                                          \\
                           & \multirow{-4}{*}{LeNet-1}   & DeepSensor                                                    & \textbf{18,747}                                                                & \textbf{3,168}                                                                      & \textbf{3.168}                                                 \\ \cline{2-6} 
                           &                             & DeepXplore                                                    & 18,474                                                                         & 2,379                                                                               & 2.379                                                          \\
                           &                             & DLFuzz                                                        & 16,809                                                                         & 1,473                                                                               & 1.473                                                          \\
                           &                             & DeepHunter                                                    & 18,420                                                                         & 2,489                                                                               & 2.489                                                          \\
                           & \multirow{-4}{*}{LeNet-4}   & DeepSensor                                                    & \textbf{18,771}                                                                & \textbf{2,868}                                                                      & \textbf{2.868}                                                 \\ \cline{2-6} 
                           &                             & DeepXplore                                                    & 18,076                                                                         & 2,224                                                                               & 2.224                                                          \\
                           &                             & DLFuzz                                                        & 16,914                                                                         & 1,509                                                                               & 1.509                                                          \\
                           &                             & DeepHunter                                                    & 18,323                                                                         & 2,469                                                                               & 2.469                                                          \\
\multirow{-12}{*}{MNIST}   & \multirow{-4}{*}{LeNet-5}   & DeepSensor                                                    & \textbf{18,349}                                                                & \textbf{2,906}                                                                      & \textbf{2.906}                                                 \\ \hline
                           &                             & DeepXplore                                                    & 16,723                                                                         & 1,162                                                                               & 1.162                                                          \\
                           &                             & DLFuzz                                                        & 15,989                                                                         & 1,336                                                                               & 1.336                                                          \\
                           &                             & DeepHunter                                                    & 17,384                                                                         & 1,549                                                                               & 1.549                                                          \\
                           & \multirow{-4}{*}{ResNet-20} & DeepSensor                                                    & \textbf{18,025}                                                                & \textbf{2,024}                                                                      & \textbf{2.024}                                                 \\ \cline{2-6} 
                           &                             & DeepXplore                                                    & 16,759                                                                         & 1,007                                                                               & 1.007                                                          \\
                           &                             & DLFuzz                                                        & 15,567                                                                         & 1,107                                                                               & 1.107                                                          \\
                           &                             & DeepHunter                                                    & 17,693                                                                         & 1,680                                                                               & 1.680                                                          \\
\multirow{-8}{*}{CIFAR-10} & \multirow{-4}{*}{VGG16}     & DeepSensor                                                    & \textbf{17,784}                                                                & \textbf{1,964}                                                                      & \textbf{1.964}                                                 \\ \hline
                           &                             & DeepXplore                                                    & 13,974                                                                         & 689                                                                                 & 0.689                                                          \\
                           &                             & DLFuzz                                                        & 14,046                                                                         & 1,974                                                                               & 1.974                                                          \\
                           &                             & DeepHunter                                                    & 15,120                                                                         & 2,287                                                                               & 2.287                                                          \\
                           & \multirow{-4}{*}{MobileNet} & DeepSensor                                                    & \textbf{15,524}                                                                & \textbf{2,994}                                                                      & \textbf{2.994}                                                 \\ \cline{2-6} 
                           &                             & DeepXplore                                                    & 11,894                                                                         & 673                                                                                 & 0.673                                                          \\
                           &                             & DLFuzz                                                        & 12,271                                                                         & 1,281                                                                               & 1.281                                                          \\
                           &                             & DeepHunter                                                    & 12,985                                                                         & 1,590                                                                               & 1.590                                                          \\
\multirow{-8}{*}{ImageNet} & \multirow{-4}{*}{ResNet-50} & DeepSensor                                                    & \textbf{13,042}                                                                & \textbf{2,010}                                                                      & \textbf{2.010}                                                 \\ \bottomrule
\end{tabular}
}

\label{adv input}
\end{table}

\textbf{Results and Analysis.} Generally speaking, DeepSensor is more effective in finding adversarial inputs than SOTA testing baselines.
DeepSensor successfully triggers more diverse errors ($\sim \times1.2$), i.e., averagely triggers more categories. For instance, on ResNet-20 of CIFAR-10 dataset, the test errors of DeepSensor is 18,025, which is 1.1 times that of DLFuzz. In addition, on ResNet-50 of ImageNet dataset, the average categories of DeepSensor is 2.010, almost 3 times and 1.3 times that of DeepXplore and DeepHunter, respectively.
The reason is that DeepSensor focuses on excitable neurons, which are related to model's loss changes. By maximizing the number of excitable neurons during iterations, DeepSensor can generate diverse testing examples that induce misclassifications. On larger models, DeepSensor exhibits stable performance in covering error categories (achieving up to 2 times of the baselines), which shows striking contract to DeepXplore's performance decrease. 

\subsubsection{Finding Errors Due to Faulty Training Procedure}
In this part, we focus on two different defects on image classification models due to faulty training: polluted data and incomplete training.

\textbf{On Polluted Data.} The results are shown in Table~\ref{polluted}. Different percentages of polluted data are taken into consideration.

\begin{table}[htbp]
\centering
\caption{Effectiveness in finding errors due to polluted data.}
\resizebox{0.7\linewidth}{!}{
\begin{tabular}{ccccccc}
\hline
\multirow{2}{*}{\textbf{Datasets}} & \multirow{2}{*}{\textbf{Models}} & \multirow{2}{*}{\textbf{$\alpha$}} & \multicolumn{4}{c}{\textbf{\#Test errors}}      \\ \cline{4-7} 
                                   &                                  &                                    & DeepXplore & DLFuzz & DeepHunter & DeepSensor   \\ \hline
\multirow{3}{*}{MNIST}             & \multirow{3}{*}{LeNet-5}         & 10\%                               & 74         & 204    & 192        & \textbf{757} \\
                                   &                                  & 20\%                               & 79         & 164    & 181        & \textbf{723} \\
                                   &                                  & 30\%                               & 72         & 172    & 185        & \textbf{784} \\ \hline
\multirow{3}{*}{CIFAR-10}          & \multirow{3}{*}{VGG16}           & 10\%                               & 64         & 91     & 101        & \textbf{590} \\
                                   &                                  & 20\%                               & 60         & 127    & 92         & \textbf{562} \\
                                   &                                  & 30\%                               & 62         & 121    & 147        & \textbf{599} \\ \hline
\multirow{3}{*}{ImageNet}          & \multirow{3}{*}{ResNet-50}       & 10\%                               & 42         & 83     & 64         & \textbf{511} \\
                                   &                                  & 20\%                               & 58         & 64     & 101        & \textbf{648} \\
                                   &                                  & 30\%                               & 50         & 61     & 77         & \textbf{639} \\ \hline
\end{tabular}
}
\label{polluted}
\end{table}

\textbf{Implementation Details.} (1) We retrain the model with manually polluted data by randomly drawing $\alpha\%$ examples, stamping perturbations of random sizes at random location, and re-labeling them as one of the original classes, i.e., the polluted class. Configuration details can be found in Table~\ref{setting}. (2) We generate 1,000 testing examples using DeepSensor and three baselines. (3) For measurement, we count test errors triggered by testing examples. 

\textbf{Results and Analysis.} As shown in Table~\ref{polluted}, DeepSensor performs quite well in finding more errors due to polluted labels, which triggers almost 5 times errors more than baselines among all datasets and models. Specifically, on LeNet-5 model with different $\alpha\%$, the total test errors found by DeepSensor is 754.7 on average, almost 10 times than that of DeepXplore. 
One of the reasons can be that the polluted and benign data jointly contribute to the model training.
For coverage-guided baselines, the benign and polluted examples may show similar neuron coverage. Thus, they become ineffective in triggering errors. On the contrary, defects placed during training are likely to be captured by excitable neurons. We speculate the possible reason that in the polluted models, neurons that will be activated by polluted perturbations contribute more to model loss changes. During fuzzing, these neurons are identified as excitable neurons. By maximizing the number of them, DeepSensor can easily hunt errors left by polluted data.

\begin{table}[htbp]
\centering
\caption{Effectiveness in finding test errors in incompletely-trained DNNs.}
\resizebox{0.75\linewidth}{!}{
\begin{tabular}{ccccccc}
\hline
\multirow{2}{*}{\textbf{Datasets}} & \multirow{2}{*}{\textbf{Models}} & \multirow{2}{*}{\textbf{Config.}} & \multicolumn{4}{c}{\textbf{\#Test errors}}      \\ \cline{4-7} 
                                   &                                  &                                   & DeepXplore & DLFuzz & DeepHunter & DeepSensor   \\ \hline
\multirow{3}{*}{MNIST}             & \multirow{3}{*}{LeNet-5}         & underfitting                      & 880        & 814    & 891        & \textbf{981} \\
                                   &                                  & well-trained                      & 894        & 845    & 906        & \textbf{937} \\
                                   &                                  & overfitting                       & 837        & 826    & 898        & \textbf{979} \\ \hline
\multirow{3}{*}{CIFAR-10}          & \multirow{3}{*}{VGG16}           & underfitting                      & 813        & 790    & 853        & \textbf{949} \\
                                   &                                  & well-trained                      & 831        & 791    & 858        & \textbf{901} \\
                                   &                                  & overfitting                       & 820        & 808    & 829        & \textbf{958} \\ \hline
\multirow{3}{*}{ImageNet}          & \multirow{3}{*}{ResNet-50}       & underfitting                      & 513        & 600    & 520        & \textbf{796} \\
                                   &                                  & well-trained                      & 525        & 604    & 627        & \textbf{672} \\
                                   &                                  & overfitting                       & 545        & 580    & 611        & \textbf{805} \\ \hline
\end{tabular}
}
\label{incomplete}
\end{table}

\textbf{On Incompletely-trained DNNs.} The experimental results are shown in Table~\ref{incomplete}, where well-trained models are still listed here for comparison.

\textbf{Implementation Details.} (1) We trained overfitting and underfitting models for LeNet-5, VGG16 and ResNet-50, and the configurations of model settings are also presented in Table~\ref{setting}. (2) We generate 1,000 testing examples to test these models and count the number of test errors. 

\textbf{Results and Analysis.} According to Table~\ref{incomplete}, DeepSensor finds more test errors, superior to all three baselines on all models, regardless of overfitting or underfitting ones. For instance, on ResNet-50, the number of test errors found by DeepSensor is 757.7 on average, which is 1.2 times that of DeepHunter.
Specifically, DeepSensor can generate more effective testing examples that hunt more errors for incompletely-trained models than for well-trained models.
On the contrary, three baselines do not show such distinct characteristics in three types of models.
DeepSensor ide
We conjecture that overfitting or underfitting DNNs are more likely to change model loss due to incomplete training. Identified by model loss changes, excitable neurons are easy to target for DeepSensor. By maximizing the number of excitable neurons, testing examples can effectively hunt these errors. 


\begin{framed}
\textbf{Answer to RQ1:} DeepSensor outperforms the SOTA methods on image classification models in two aspects: (1) finding more adversarial inputs than baselines ($\sim\times1.2$ on average); (2) finding more test errors due to faulty training: on polluted data ($\sim\times5$ on average) and incompletely-trained models ($\sim\times1.2$ on average). 
\end{framed}

\subsection{RQ2: Robustness Improvement}
When answering this question, we refer to adversarial robustness improvement measured by empirical measurement ASR and certified robustness metric CLEVER~\cite{weng2018evaluating}.

The robustness improvement highly depends on the quality of the retraining data, i.e., generated testing examples. By retraining the model with the generated examples, the robustness of the model can be improved significantly. In the experiment, we retrain the model using 500 benign examples and their 500 corresponding testing examples by DeepSensor and baselines for 20 epochs with $batch\_size$=32.  

PGD-based adversarial training (PGD-AT) \cite{Madry2018Towards} has been demonstrated to be the only method that can train moderately robust DNNs without being fully attacked \cite{athalye2018obfuscated}. So we also compare the robustness of retrained model by DeepSensor and PGD-AT. For retraining parameters, 500 benign examples and 500 PGD adversarial examples are used, with 20 epochs and $batch\_size$=32. Noted that the benign accuracy of all retrained models is barely affected by the retraining.

\subsubsection{Improvement Measured by ASR} 
\begin{figure}[htbp]
\centering
        \includegraphics[width=0.5\linewidth]{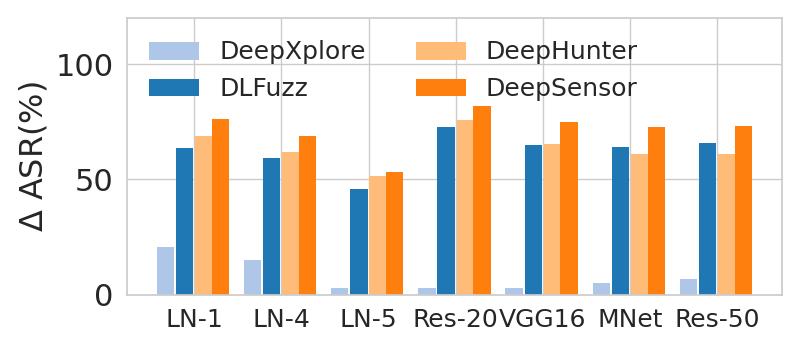}
\caption{$\Delta$ASR between original models and those retrained with testing examples.}
	\label{withtesting}
\end{figure}

\begin{figure}[htbp]
\centering
  \subfigure[LeNet-5]{
  \includegraphics[width=0.28\linewidth]{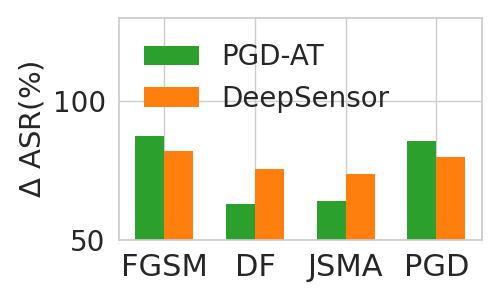} }
  \hspace{-10pt}
  \subfigure[VGG16]{
  \includegraphics[width=0.28\linewidth]{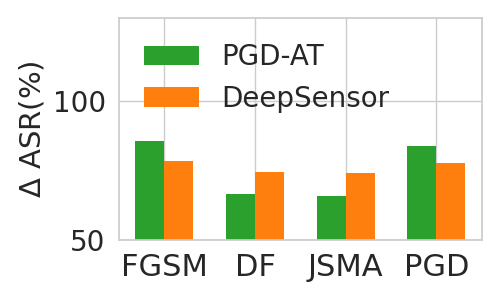} }
  \hspace{-10pt}
  \subfigure[ResNet-50]{
  \includegraphics[width=0.28\linewidth]{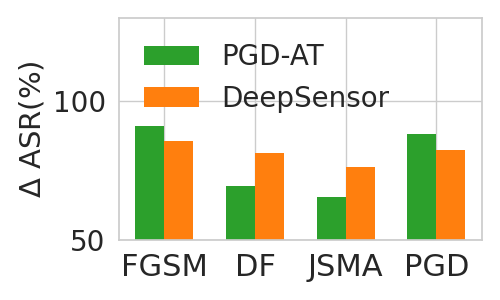} }
\caption{$\Delta$ASR between original models and PGD-AT and those with testing examples of DeepSensor. DeepFool is abbreviated as ``DF''.}
 \label{withadversarial}
\end{figure}

The result is shown in Fig. \ref{withtesting} and \ref{withadversarial}, where the former represents retraining with testing examples while the latter represents that with PGD adversarial examples.

\textbf{Implementation Details.} 
(1) For evaluating retraining models with testing examples, we attack the retrained model using the SOTA first-order attack PGD~\cite{Madry2018Towards}. $\epsilon$ in PGD is set to 0.3 for MNIST and CIFAR-10 and 0.5 for ImageNet. 1,000 adversarial examples are generated for evaluation. 
(2) For evaluating retraining models with PGD-AT, we retrain the model with PGD-AT by 500 PGD adversarial examples. For comparison, we attack the retrained model using fast gradient sign method (FGSM)~\cite{GoodfellowSS14}, DeepFool~\cite{moosavi2016deepfool}, Jacobian-based saliency map attack (JSMA)~\cite{papernot2016limitations} and PGD. 1,000 adversarial examples per attack are adopted. 
(3) For measurement, we compare the ASR of the retrained model and the original model and calculate the decrease in ASR, denoted as $\Delta$ASR. The larger, the better. 


\textbf{Results and Analysis.} Compared with baselines, models retrained with testing examples generated by DeepSensor are more robust against PGD attack. As shown in Fig.\ref{withtesting}, bars of orange are the highest among all models and methods. This indicates DeepSensor can cause more decrease in ASR, which in turn demonstrates the better quality of testing examples generated by it.

Compared with models after PGD-AT, in all cases, models trained with testing examples generated by DeepSensor doesn't show inferior robustness results. Specifically, the robustness against DeepFool and JSMA are higher than that by PGD-AT. In Fig. \ref{withadversarial}, $\Delta$ASR after retraining by DeepSensor are all over 70\%. We speculate that models after PGD-AT are more robust against gradient-based attacks like FGSM. Since DeepSensor takes diverse neurons into account, models retrained by testing examples generated by it show general robustness against various attacks. 

\subsubsection{Improvement Measured by CLEVER}
\begin{figure}[htbp]
\centering
        \includegraphics[width=0.55\linewidth]{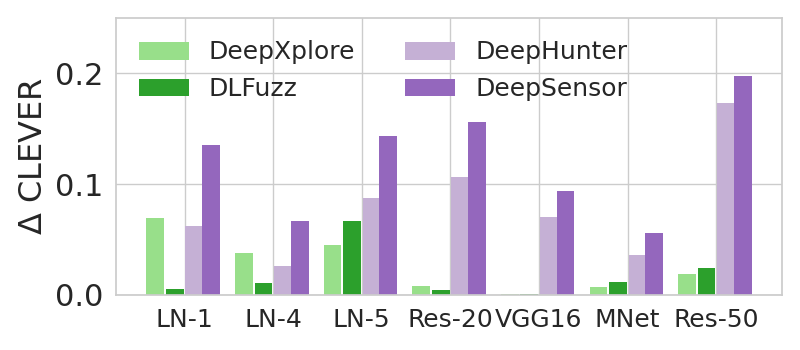}
\caption{$\Delta$CLEVER between original and retrained models with testing examples.}
	\label{clever}
\end{figure}

The result is shown in Fig. \ref{clever}, where different bars denote the improvement of CLEVER after retraining.

\textbf{Implementation Details.} (1) We measure the improvement in CLEVER, denoted as $\Delta$CLEVER, after retraining using testing examples generated by DeepSensor and baselines. The larger, the better. 
(2) The sampling parameters $N_b$ and $N_s$ in CLEVER are set to 500 and 1024, respectively.

\textbf{Results and Analysis.} CLEVER scores on retrained models are significantly improved by DeepSensor. As can be seen, bars of deep purple are the highest among all bars, which indicates that DeepSensor can help build more robust models than baselines. DeepSensor increases certified robustness by 0.12 on average, almost 3 times better than all baselines. It is mainly because DeepSensor targets neurons calculated from model loss, which is closely related to model robustness. So larger adversarial perturbations are required to fool the retrained models. Thus, CLEVER scores of DeepSensor are larger.

\begin{framed}
\textbf{Answer to RQ2:} DeepSensor is superior to testing baselines in robustness improvement through retraining in two aspects: (1) leads to more decrease ($\sim\times1.6$) on ASR; (2) achieves larger CLEVER scores ($\sim\times3$) than baselines on average.  
\end{framed}

\subsection{RQ3: Interpretation}
When interpreting effective testing results, we refer to excitable neurons and testing visualization. 

\subsubsection{Excitable Neurons And Adversarial Examples}
\begin{figure}[htbp]
\centering
        \includegraphics[width=0.7\linewidth]{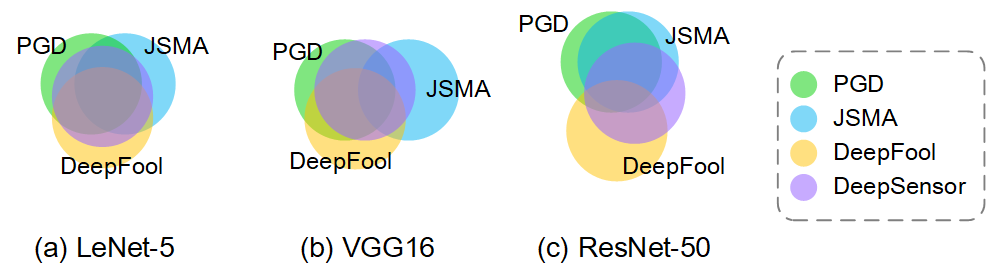}
\caption{The overlap of excitable neurons between testing examples of DeepSensor and adversarial examples.}
	\label{iou}
\end{figure}

We further study the relationship between adversarial examples and testing examples generated by DeepSensor. The overlap of excitable neurons between DeepSensor and adversarial examples is shown in Fig. \ref{iou}. 

\textbf{Implementation Details.} (1) We randomly select 100 adversarial examples from three different attacks: PGD, JSMA and DeepFool. 100 testing examples generated by DeepSensor after 20 iterations are also used. (2) We calculate the overlap of $top$-10 excitable neurons. (3) In Fig. \ref{iou}, green, blue, yellow and purple circles denote excitable neurons activated by PGD, JSMA, DeepFool and DeepSensor, respectively. Each circle includes 10 neurons in total. The overlapped area indicates the number of overlapped neurons on average.

\textbf{Results and Analysis.} We observe that excitable neurons activated by DeepSensor can cover almost all those activated by three adversarial attacks, e.g., purple circle overlaps green and yellow circles on LeNet-5 and VGG16. Three circles of adversarial examples are relatively scattered from each other. This shows that although some neurons are frequently and repeatedly activated by different attacks, in general, different attacks exploit different neurons. Adversarial attacks are designed to fool the model, so they only search for a few neurons that can directly lead to misclassifications. On the contrary, DeepSensor takes frequently-used neurons by multiple attacks into consideration so it can trigger test errors from various categories. 


\subsubsection{Visualizations}
In this part, we interpret testing effectiveness through visualizations of t-SNE~\cite{van2008visualizing} and testing examples.

\begin{figure}[htbp]
  \centering
\includegraphics[width=0.6\linewidth]{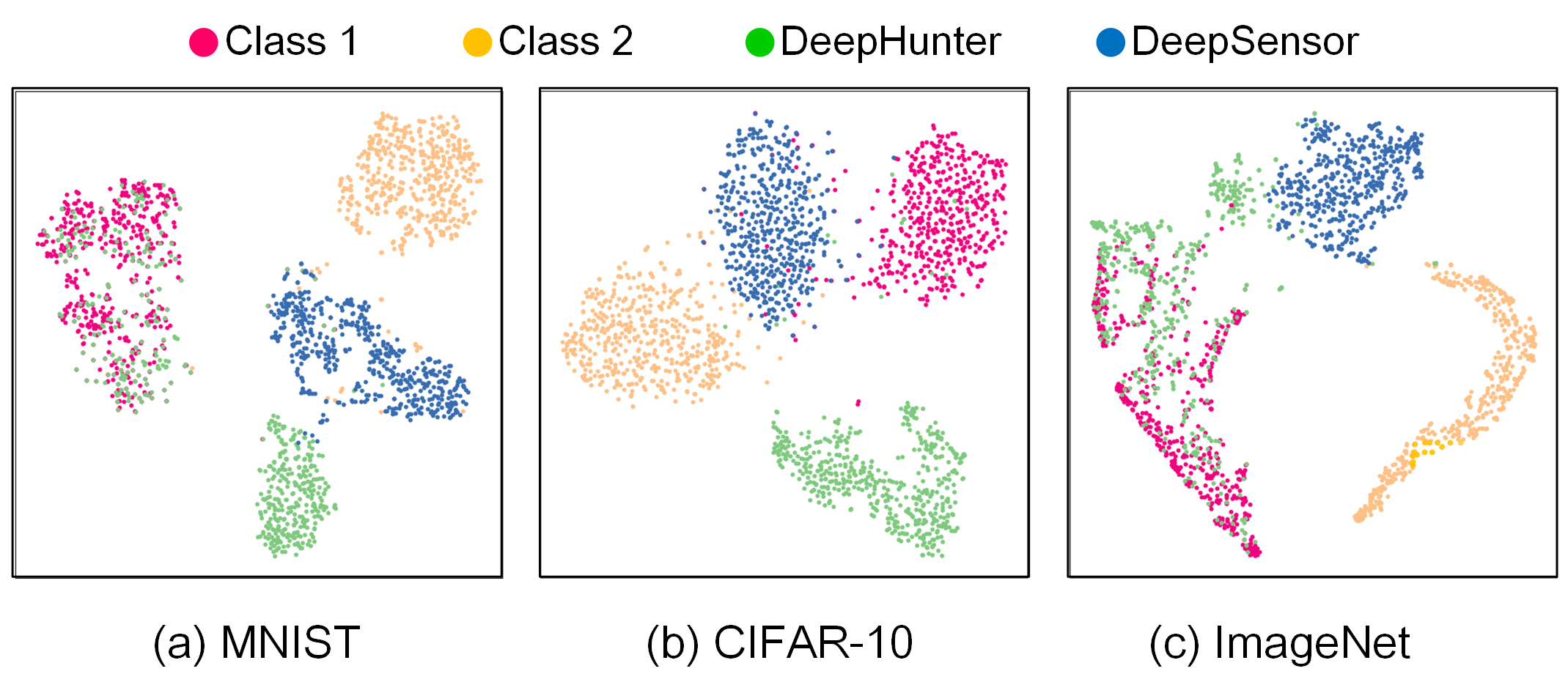}
  \caption{t-NSE visualizations on MNIST, CIFAR-10 and ImageNet.}
  \label{fig:tsne}
\end{figure}

\textbf{Visualization of t-SNE.} We visualize the high-dimensional representation of benign examples and the generated testing examples via t-SNE, as shown in Fig.~\ref{fig:tsne}.

\textbf{Implementation Details.} (1) We visualize two classes of benign examples (``Class 1'' and ``Class 2''), and the testing examples generated by DeepHunter and our DeepSensor using ``Class 1'' examples as benign seeds. DeepHunter is chosen as baseline here due to its superior testing performance. (2) LeNet-5 of MNIST, ResNet-20 of CIFAR-10 and ResNet-50 of ImageNet are adopted for visualizations. (3) We only visualize the generated examples that are classified as ``Class 1'' and ``Class 2'' for a clear demonstration.

\textbf{Results and Analysis.} DeepHunter and DeepSensor can both trigger corner cases that lead to misclassification. But DeepSensor shows superior results. In Fig.~\ref{fig:tsne}, the testing examples of DeepHunter and DeepSensor form two clusters different from the two benign classes, which indicates both of them can find errors leading to misclassifications. However, the testing examples generated by DeepHunter considerably overlap with ``Class 1''. It means that these examples may not be able to cause test errors. This also explains the experimental results in Table~\ref{adv input} that DeepHunter triggers fewer test errors than DeepSensor. On the contrary, blue clusters are separated from other ones, which demonstrates better error-triggering capability of DeepSensor.

\begin{figure}[htbp]
\renewcommand{\@thesubfigure}{\hskip\subfiglabelskip}
\makeatother
  \centering
  \subfigure[``6'']{
  \includegraphics[width=0.08\linewidth]{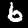} }
  \subfigure[``0'']{
  \includegraphics[width=0.08\linewidth]{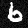} }
  \subfigure[$\rho \times10$]{
  \includegraphics[width=0.08\linewidth]{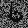} }
  \subfigure[``7'']{
  \includegraphics[width=0.08\linewidth]{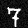} }
  \subfigure[``2'']{
  \includegraphics[width=0.08\linewidth]{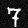} }
 \subfigure[$\rho \times10$]{
  \includegraphics[width=0.08\linewidth]{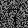} }  
    \subfigure[``9'']{
  \includegraphics[width=0.08\linewidth]{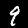} }
  \subfigure[``1'']{
  \includegraphics[width=0.08\linewidth]{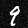} }
  \subfigure[$\rho \times10$]{
  \includegraphics[width=0.08\linewidth]{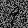} }\\
  \subfigure[``plane'']{
  \includegraphics[width=0.08\linewidth]{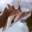} }
  \subfigure[``frog'']{
  \includegraphics[width=0.08\linewidth]{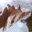} }
  \subfigure[$\rho \times10$]{
  \includegraphics[width=0.08\linewidth]{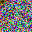} }
    \subfigure[``car'']{
  \includegraphics[width=0.08\linewidth]{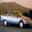} }
  \subfigure[``horse'']{
  \includegraphics[width=0.08\linewidth]{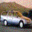} }
  \subfigure[$\rho \times10$]{
  \includegraphics[width=0.08\linewidth]{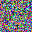} } 
    \subfigure[``Ship'']{
  \includegraphics[width=0.08\linewidth]{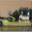} }
  \subfigure[``Truck'']{
  \includegraphics[width=0.08\linewidth]{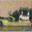} }
  \subfigure[$\rho \times10$]{
  \includegraphics[width=0.08\linewidth]{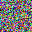} }\\
  \subfigure[``egret'']{
  \includegraphics[width=0.08\linewidth]{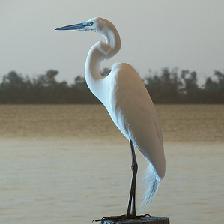} }
  \subfigure[``velvet'']{
  \includegraphics[width=0.08\linewidth]{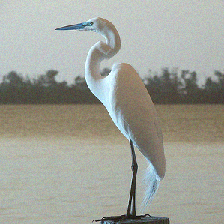} }
  \subfigure[$\rho \times10$]{
  \includegraphics[width=0.08\linewidth]{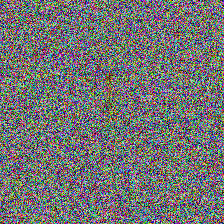} }
  \subfigure[``wolf'']{
  \includegraphics[width=0.08\linewidth]{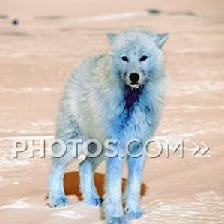} }  
  \subfigure[``tabby'']{
  \includegraphics[width=0.08\linewidth]{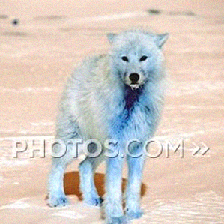} }  
  \subfigure[$\rho \times10$]{
  \includegraphics[width=0.08\linewidth]{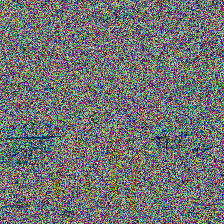} }
  \subfigure[``Fox'']{
  \includegraphics[width=0.08\linewidth]{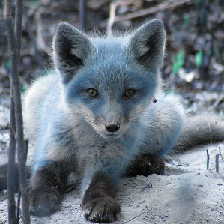} }
  \subfigure[``corn'']{
  \includegraphics[width=0.08\linewidth]{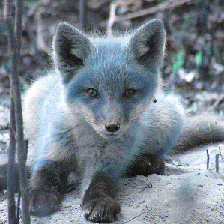} }
  \subfigure[$\rho \times10$]{
  \includegraphics[width=0.08\linewidth]{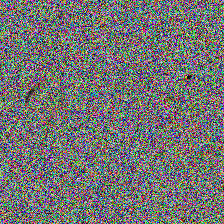} }\\

  \caption{Testing examples on MNIST, CIFAR-10 and ImageNet datasets generated by DeepSensor.}
  \label{fig:test_exp}
\end{figure}

\textbf{Visualization of Testing Examples}. The results are shown in Fig.~\ref{fig:test_exp}.

\textbf{Implementation Details.} (1) The first, middle and last row are from MNIST, CIFAR-10 and ImageNet datasets, respectively. Columns 1 and 4 show original images with their labels attached at the bottom of them, while columns 2 and 5 are testing examples with their labels. (2) We also visualize the perturbations in columns 3 and 6, denoted as $\rho$ in the caption. They are all magnified 10 times for better visualization.

\makeatletter

\textbf{Results and Analysis.} It can be easily observed that testing examples closely resemble original images, with perturbation nearly imperceptible to human eyes. This well demonstrates the performance of DeepSensor that corner cases could be triggered by small size of perturbations.

\begin{framed}
\textbf{Answer to RQ3:} The main reason of effective testing goes to that (1) DeepSensor uses diverse neurons, which is beneficial to the diversity of testing examples; (2) the more separated clusters with benign classes in high-dimensional feature distribution.
\end{framed}

\subsection{RQ4: Application}
Speaker recognition systems are widely used in the real world. We also conduct experiments of DeepSensor on speaker recognition models to study its generality among scenarios. The results are shown in Table \ref{adv_speaker} and \ref{incomplete_speaker}. 

\textbf{Implementation Details.} (1) We run DeepSensor and baselines using 4,000 benign seeds for 20 iterations. (2) For measurement, we count the total numbers of test errors due to adversarial inputs and incomplete training. Since models trained with polluted data suffer from large decrease of benign accuracy, we don't conduct testing on scenario of polluted data. (3) Model configurations are shown in Table~\ref{data_model}. For incomplete-trained models, $batch\_size$ of Deep Speaker is set to 20. For the underfitting model, epoch is 5 with its acc=71.16\%. For the overfitting model, epoch is 80 with its acc=76.26\%.

\begin{table}[htbp]
\centering
\caption{Adversarial inputs found on VCTK datasets.}
\resizebox{0.8\linewidth}{!}{
\begin{tabular}{cccccc}
\toprule
\textbf{Datasets}                   & \textbf{Models }                     & \begin{tabular}[c]{@{}c@{}} \textbf{Methods}  \end{tabular} & { \begin{tabular}[c]{@{}c@{}}  \textbf{\#Test} \textbf{errors} \end{tabular}} & { \begin{tabular}[c]{@{}c@{}}\textbf{\#Error } \textbf{categories}\end{tabular}} & \begin{tabular}[c]{@{}c@{}}\textbf{\#Average } \textbf{categories} \end{tabular} \\ \hline 
&  & DeepXplore & 15,716 & 2,766 & 2.766 \\
 &  & DLFuzz & 16,962 & 3,194 & 3.194 \\
 &  & DeepHunter & 17,314 & 3,065 & 3.065 \\
 &
  \multirow{-4}{*}{ResNet-34} &
  DeepSensor &
  \textbf{18,550} &
  \textbf{3,997} &
  \textbf{3.997} \\ \cline{2-6} 
 &  & DeepXplore & 14,793 & 3,012 & 3.012 \\
 &  & DLFuzz & 16,034 & 3,721 & 3.721 \\
 &  & DeepHunter & 16,787 & 3,684 & 3.684 \\
\multirow{-8}{*}{VCTK} &
  \multirow{-4}{*}{Deep Speaker} &
  DeepSensor &
  \textbf{18,016} &
  \textbf{4,250} &
  \textbf{4.250} 
\\ \bottomrule
\end{tabular}
}
\label{adv_speaker}
\end{table}

\begin{table}[htbp]
\centering
\caption{Test errors found in incompletely-trained DNNs on VCTK dataset.}
\resizebox{0.65\linewidth}{!}{
\begin{tabular}{cccccc}
\toprule
\multirow{2}{*}{\textbf{Models} }    & \multirow{2}{*}{\textbf{Config.} } & \multicolumn{4}{c}{\textbf{\#Test errors} }               \\ \cline{3-6} 
                           &                          & DeepXplore & DLFuzz & DeepHunter & DeepSensor   \\ \hline
                           \multirow{3}{*}{Deep Speaker} & underfitting     & 762        & 815    & 810        & \textbf{983}        \\
                              & well-trained     & 739        & 811    & 827        & \textbf{900}        \\
                              & overfitting      & 744        & 800    & 823        & \textbf{967}        \\ \bottomrule
\end{tabular}
}
\label{incomplete_speaker}
\end{table}

\textbf{Results and Analysis.} It can be observed from Table \ref{adv_speaker} and \ref{incomplete_speaker} that DeepSensor can explore more defects due to different causes than testing baselines. For instance, on Deep Speaker, DeepSensor can find 1.2 times more adversarial inputs that belong to 1.4 times more categories than DeepXplore. Besides, the number of test errors found by DeepSensor due to incomplete training is 1.2 times on average than that of baselines. This demonstrates the superiority of DeepSensor. By maximizing the number of excitable neurons, more test errors from diverse categories can be triggered.

\begin{framed}
\textbf{Answer to RQ4:} Apart from image classification models, DeepSensor can be applied to test speaker recognition models with superior performance: finding more adversarial inputs ($\sim$ $\times$1.2 on average) and more errors in incompletely-trained models ($\sim$ $\times$1.2).
\end{framed}

\subsection{RQ5: Parameter Sensitivity}
When reporting the results, we refer to the impact of different parameters and PSO initialization strategy.

\subsubsection{Influence of $\lambda$}
\begin{figure}[htbp]
\centering
\subfigure[Impact of $\lambda$]{
\label{fig:lamda}
  \includegraphics[width=0.3\linewidth]{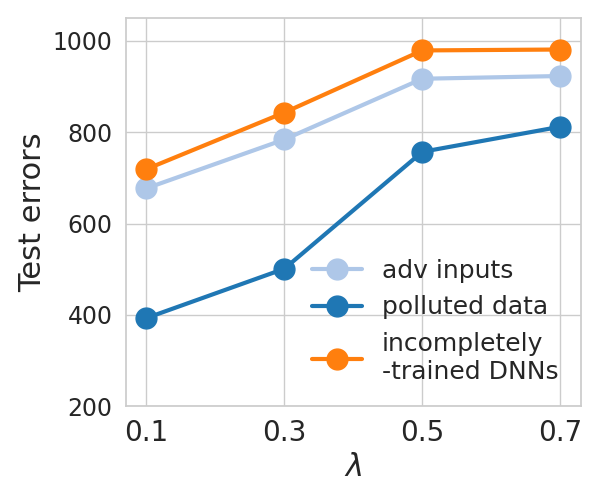} }
\subfigure[Impact of $\epsilon$]{
\label{fig:epsilon}
  \includegraphics[width=0.3\linewidth]{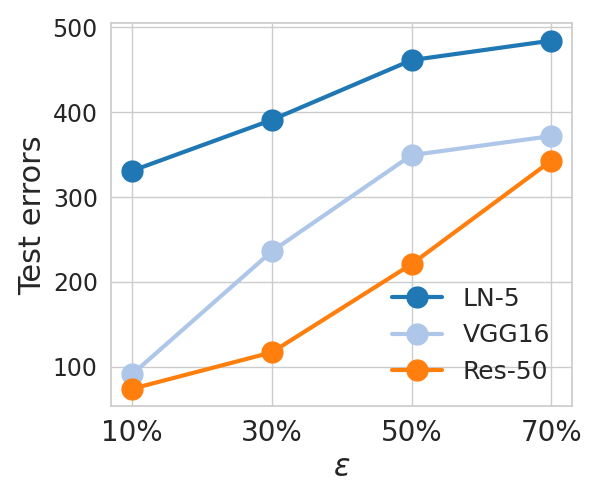} }
\caption{Impact of excitable neurons on DeepSensor.}
\label{ss_para2}
\end{figure}

$\lambda$ defines what neurons are considered as excitable neurons in DeepSensor, the value of which is crucial to DeepSensor's performance. The impact of $\lambda$ is shown in Fig.~\ref{fig:lamda}.

\textbf{Implementation Details.} (1) We use 1000 testing examples for LeNet-5 of MNIST model and investigate the impact of different $\lambda$. $\alpha$ in polluted data is 10\% and overfitting model is used. (2) For measurement, the number of test errors from adversarial inputs, polluted data and incompletely-trained models are counted.  

\textbf{Results and Analysis.} With the increase of $\lambda$, the number of test errors found by DeepSensor gradually increases and then remains stable after 0.6.
A smaller $\lambda$ means that the neurons DeepSensor focuses on during the optimization are less excitable.
As a consequence, the testing examples generated may not benefit from the ``true'' excitable neurons, and it benefits DeepSensor by increasing $\lambda$.
However, a larger $\lambda$ means the neurons with larger Shapley value are focused by DeepSensor during the optimization and fewer excitable neurons are considered in the process, which hence limits the optimization objective, i.e., maximizing the number of excitable neurons. 

\subsubsection{Influence of $\epsilon$} 
$\epsilon$ controls the percentage of excitable neurons in the fitness function during iterations. The impact of $\epsilon$ is shown in Fig.~\ref{fig:epsilon}.

\textbf{Implementation Details.} (1) 500 testing examples are used for evaluation. LeNet-5 of MNIST, VGG16 of CIFAR-10 and ResNet-50 are adopted. (2) For measurement, the numbers of test errors due to adversarial inputs under different $\epsilon$ are calculated.

\textbf{Results and Analysis.} As the figure suggests, the number of test errors found by DeepSensor increases with the increase of the percentage of excitable neurons. We assume that it is more likely to expose erroneous behaviors with more excitable neurons activated. The result well supports our assumption. By maximizing the number of excitable neurons, more corner case behaviors can be triggered by DeepSensor.

\subsubsection{Influence of Random Initialization Strategy in PSO\label{pso_random}}
\begin{figure}[htbp]
\centering
\subfigure[LeNet-5]{
  \includegraphics[width=0.35\linewidth]{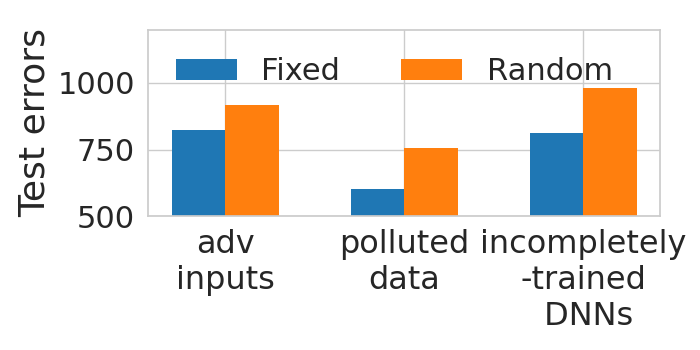} }
  \subfigure[ResNet-50]{
  \includegraphics[width=0.35\linewidth]{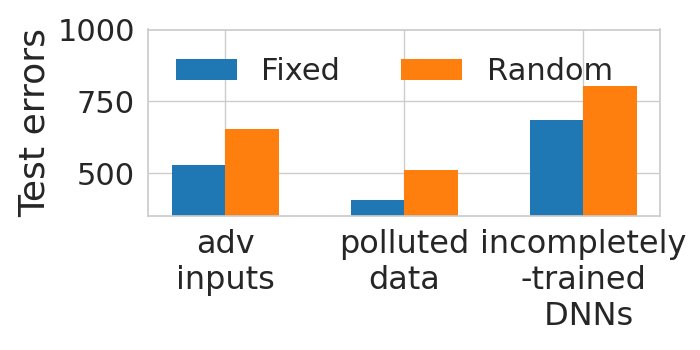} }
\caption{The impact of random location initialization.}
\label{random}
\end{figure}

Here we analyze the effectiveness of random initialization strategy in DeepSensor, which is proposed for avoiding local optimum. Results on LeNet-5 and ResNet-50 are shown in Fig. \ref{random}.

\textbf{Implementation Details.} (1) We compare random location initialization (i.e., the initial location of each swarm before iterations is randomly chosen) with the fixed one. (2) 1,000 testing examples are generated for evaluation. For measurement, the number of test errors due to adversarial inputs, polluted data and incompletely-trained DNNs are calculated. $\alpha$ in polluted data is 10\% and overfitting model is used.

\textbf{Results and Analysis.} It can be easily observed that test errors of random strategy outperforms that of fixed one on two models. In Fig. \ref{random}, orange bars are all higher than blue ones. This indicates that more excitable neurons can be found during iterations when particle position is random initialized. As a result, more test errors can be triggered regardless of their reasons, as well as global optimal, can be reached with less time to converge. In this way, the problem of local optimum can be alleviated.

\begin{framed}
\textbf{Answer to RQ5:} DeepSensor is robust to different parameters. Besides, the random initialization of PSO also contributes to DeepSensor's effectiveness. 
\end{framed}

\section{Limitations and Discussions}
Although DeepSensor has demonstrated its effectiveness of finding diverse potential flaws inside the model, it can be further improved in the following aspects in the future. 

\textbf{Higher Efficiency.} Although DeepSensor exceeds SOTA testing techniques in the overall datasets, the efficiency of it needs to be accelerated. For instance, to generate 100 testing examples, DeepSensor takes 259.7s on LeNet-5, 1081.4s on VGG16, and 1377.3s on ResNet-50. It is much longer than the average time of baselines (i.e., 167.8s on LeNet-5, 889.1s on VGG16, and 1084.4s on ResNet-50). The operation of searching excitable neurons in the whole model may be time-consuming, beacuse each neuron needs to be calculated $N!$ times to calculate Shapley value. $N$ denotes the total number of neurons in the model. With the growth of model complexity, the computation time will increase dramatically.
So in the future, we will work towards lower algorithm complexity by targeting excitable neurons more straightforward.

\textbf{Other Optimization Solutions.} PSO is adopted by DeepSensor during iterations for its low time complexity. Other existing optimization algorithms, i.e., genetic algorithm, ant colony optimization, are widely-used for their easy implementation and generality. We have combined excitable neurons and other optimization algorithms, and further conducted experiments to study their effectiveness. Results show that test errors found are not sensitive to optimization solutions. Considering the time complexity, PSO is still an ideal choice for DeepSensor.

\textbf{Distinguish Misclassified Natural Inputs.} Detecting misclassification of natural inputs is another side of defect-finding capability. So we intend to improve DeepSensor so that it can distinguish errors due to misclassified natural inputs, which will be beneficial to a more reliable fault-detection capability.

\textbf{Applied to Different Neural Network Architectures.} In this paper, we evaluate DeepSensor on classification tasks on CNN-based models. In the future, we are committed to widely applying DeepSensor to other generation tasks (e.g., regression tasks, etc.), and more complex neural network architectures (e.g., recurrent neural networks).

\section{Conclusion}
In this paper, we attribute the cause of erroneous behaviors to neurons, and adopt Shapley value to quantitatively measure neuron's contribution towards model prediction changes due to input perturbations. Motivated by it, we propose the concept of excitable neurons and design DeepSensor, a novel white-box DNN testing approach. By maximizing the number of excitable neurons, DeepSensor can generate testing examples that perform better in finding errors and robustness improvement than SOTA testing methods. It can also be applied to test speaker recognition models. Besides, we have also provide interpretable insights between excitable neurons and testing performance for better understanding.

\section*{Acknowledgments}
This research was supported by the National Natural Science Foundation of China (No. 62072406), the Natural Science Foundation of Zhejiang Province (Nos. LY19F020025 and LGF20F020016), the Key R\&D Projects in Zhejiang Province (Nos. 2021C01117 and 2022C01018).


\begin{thebibliography}{10}
\providecommand{\url}[1]{#1}
\csname url@samestyle\endcsname
\providecommand{\newblock}{\relax}
\providecommand{\bibinfo}[2]{#2}
\providecommand{\BIBentrySTDinterwordspacing}{\spaceskip=0pt\relax}
\providecommand{\BIBentryALTinterwordstretchfactor}{4}
\providecommand{\BIBentryALTinterwordspacing}{\spaceskip=\fontdimen2\font plus
\BIBentryALTinterwordstretchfactor\fontdimen3\font minus
  \fontdimen4\font\relax}
\providecommand{\BIBforeignlanguage}[2]{{%
\expandafter\ifx\csname l@#1\endcsname\relax
\typeout{** WARNING: IEEEtran.bst: No hyphenation pattern has been}%
\typeout{** loaded for the language `#1'. Using the pattern for}%
\typeout{** the default language instead.}%
\else
\language=\csname l@#1\endcsname
\fi
#2}}
\providecommand{\BIBdecl}{\relax}
\BIBdecl

\bibitem{zheng2019denoising}
Y.~Zheng, H.~Duan, X.~Tang, C.~Wang, and J.~Zhou, ``Denoising in the dark:
  Privacy-preserving deep neural network-based image denoising,'' \emph{IEEE
  Transactions on Dependable and Secure Computing}, vol.~18, no.~3, pp.
  1261--1275, 2019.

\bibitem{liang2018detecting}
B.~Liang, H.~Li, M.~Su, X.~Li, W.~Shi, and X.~Wang, ``Detecting adversarial
  image examples in deep neural networks with adaptive noise reduction,''
  \emph{IEEE Transactions on Dependable and Secure Computing}, pp. 1--14, 2018.

\bibitem{AgarwalSVR21}
A.~Agarwal, R.~Singh, M.~Vatsa, and N.~K. Ratha, ``Image transformation-based
  defense against adversarial perturbation on deep learning models,''
  \emph{{IEEE} Trans. Dependable Secur. Comput.}, vol.~18, no.~5, pp.
  2106--2121, 2021.

\bibitem{ma2018efficient}
H.~Ma, R.~Zhang, G.~Yang, Z.~Song, K.~He, and Y.~Xiao, ``Efficient fine-grained
  data sharing mechanism for electronic medical record systems with mobile
  devices,'' \emph{IEEE Transactions on Dependable and Secure Computing},
  vol.~17, no.~5, pp. 1026--1038, 2018.

\bibitem{QianDHCJL21}
J.~Qian, H.~Du, J.~Hou, L.~Chen, T.~Jung, and X.~Li, ``Speech sanitizer: Speech
  content desensitization and voice anonymization,'' \emph{{IEEE} Trans.
  Dependable Secur. Comput.}, vol.~18, no.~6, pp. 2631--2642, 2021.

\bibitem{LuLHL20}
L.~Lu, L.~Liu, M.~J. Hussain, and Y.~Liu, ``I sense you by breath: Speaker
  recognition via breath biometrics,'' \emph{{IEEE} Trans. Dependable Secur.
  Comput.}, vol.~17, no.~2, pp. 306--319, 2020.

\bibitem{lambert2016understanding}
F.~Lambert, ``Understanding the fatal tesla accident on autopilot and the nhtsa
  probe,'' \emph{Electrek, July}, vol.~1, p.~1, 2016.

\bibitem{wicker2018feature}
M.~Wicker, X.~Huang, and M.~Kwiatkowska, ``Feature-guided black-box safety
  testing of deep neural networks,'' in \emph{International Conference on Tools
  and Algorithms for the Construction and Analysis of Systems}.\hskip 1em plus
  0.5em minus 0.4em\relax Springer, 2018, pp. 408--426.

\bibitem{ma2018deepmutation}
L.~Ma, F.~Zhang, J.~Sun, M.~Xue, B.~Li, F.~Juefei-Xu, C.~Xie, L.~Li, Y.~Liu,
  J.~Zhao \emph{et~al.}, ``Deepmutation: Mutation testing of deep learning
  systems,'' in \emph{2018 IEEE 29th International Symposium on Software
  Reliability Engineering (ISSRE)}.\hskip 1em plus 0.5em minus 0.4em\relax
  IEEE, 2018, pp. 100--111.

\bibitem{pei2017deepxplore}
K.~Pei, Y.~Cao, J.~Yang, and S.~Jana, ``Deepxplore: Automated whitebox testing
  of deep learning systems,'' in \emph{proceedings of the 26th Symposium on
  Operating Systems Principles}, 2017, pp. 1--18.

\bibitem{guo2018dlfuzz}
J.~Guo, Y.~Jiang, Y.~Zhao, Q.~Chen, and J.~Sun, ``Dlfuzz: Differential fuzzing
  testing of deep learning systems,'' in \emph{Proceedings of the 2018 26th ACM
  Joint Meeting on European Software Engineering Conference and Symposium on
  the Foundations of Software Engineering}, 2018, pp. 739--743.

\bibitem{xie2019deephunter}
X.~Xie, L.~Ma, F.~Juefei-Xu, M.~Xue, H.~Chen, Y.~Liu, J.~Zhao, B.~Li, J.~Yin,
  and S.~See, ``Deephunter: a coverage-guided fuzz testing framework for deep
  neural networks,'' in \emph{Proceedings of the 28th ACM SIGSOFT International
  Symposium on Software Testing and Analysis}, 2019, pp. 146--157.

\bibitem{lee2020effective}
S.~Lee, S.~Cha, D.~Lee, and H.~Oh, ``Effective white-box testing of deep neural
  networks with adaptive neuron-selection strategy,'' in \emph{Proceedings of
  the 29th ACM SIGSOFT International Symposium on Software Testing and
  Analysis}, 2020, pp. 165--176.

\bibitem{li2019structural}
Z.~Li, X.~Ma, C.~Xu, and C.~Cao, ``Structural coverage criteria for neural
  networks could be misleading,'' in \emph{2019 IEEE/ACM 41st International
  Conference on Software Engineering: New Ideas and Emerging Results
  (ICSE-NIER)}.\hskip 1em plus 0.5em minus 0.4em\relax IEEE, 2019, pp. 89--92.

\bibitem{harel2020neuron}
F.~Harel-Canada, L.~Wang, M.~A. Gulzar, Q.~Gu, and M.~Kim, ``Is neuron coverage
  a meaningful measure for testing deep neural networks?'' in \emph{Proceedings
  of the 28th ACM Joint Meeting on European Software Engineering Conference and
  Symposium on the Foundations of Software Engineering}, 2020, pp. 851--862.

\bibitem{dong2019there}
Y.~Dong, P.~Zhang, J.~Wang, S.~Liu, J.~Sun, J.~Hao, X.~Wang, L.~Wang, J.~S.
  Dong, and D.~Ting, ``There is limited correlation between coverage and
  robustness for deep neural networks,'' \emph{arXiv preprint
  arXiv:1911.05904}, pp. 1--12, 2019.

\bibitem{shapley1953value}
L.~S. Shapley, ``A value for n-person games,'' \emph{Contributions to the
  Theory of Games}, 1953.

\bibitem{ChenSWJ19}
\BIBentryALTinterwordspacing
J.~Chen, L.~Song, M.~J. Wainwright, and M.~I. Jordan, ``L-shapley and
  c-shapley: Efficient model interpretation for structured data,'' in \emph{7th
  International Conference on Learning Representations, {ICLR} 2019, New
  Orleans, LA, USA, May 6-9, 2019}.\hskip 1em plus 0.5em minus 0.4em\relax
  OpenReview.net, 2019. [Online]. Available:
  \url{https://openreview.net/forum?id=S1E3Ko09F7}
\BIBentrySTDinterwordspacing

\bibitem{LiKLCZSX21}
\BIBentryALTinterwordspacing
J.~Li, K.~Kuang, L.~Li, L.~Chen, S.~Zhang, J.~Shao, and J.~Xiao,
  ``Instance-wise or class-wise? {A} tale of neighbor shapley for concept-based
  explanation,'' in \emph{{MM} '21: {ACM} Multimedia Conference, Virtual Event,
  China, October 20 - 24, 2021}, H.~T. Shen, Y.~Zhuang, J.~R. Smith, Y.~Yang,
  P.~Cesar, F.~Metze, and B.~Prabhakaran, Eds.\hskip 1em plus 0.5em minus
  0.4em\relax {ACM}, 2021, pp. 3664--3672. [Online]. Available:
  \url{https://doi.org/10.1145/3474085.3475337}
\BIBentrySTDinterwordspacing

\bibitem{guan2022few}
J.~Guan, Z.~Tu, R.~He, and D.~Tao, ``Few-shot backdoor defense using shapley
  estimation,'' in \emph{Proceedings of the IEEE/CVF Conference on Computer
  Vision and Pattern Recognition}, 2022, pp. 13\,358--13\,367.

\bibitem{lecun2015lenet}
Y.~LeCun \emph{et~al.}, ``Lenet-5, convolutional neural networks,'' \emph{URL:
  http://yann. lecun. com/exdb/lenet}, vol.~20, no.~5, p.~14, 2015.

\bibitem{lecun1998gradient}
Y.~LeCun, L.~Bottou, Y.~Bengio, and P.~Haffner, ``Gradient-based learning
  applied to document recognition,'' \emph{Proceedings of the IEEE}, vol.~86,
  no.~11, pp. 2278--2324, 1998.

\bibitem{van2008visualizing}
L.~Van~der Maaten and G.~Hinton, ``Visualizing data using t-sne.''
  \emph{Journal of machine learning research}, vol.~9, no.~11, pp. 1--27, 2008.

\bibitem{eberhart1995new}
R.~Eberhart and J.~Kennedy, ``A new optimizer using particle swarm theory,'' in
  \emph{MHS'95. Proceedings of the Sixth International Symposium on Micro
  Machine and Human Science}.\hskip 1em plus 0.5em minus 0.4em\relax Ieee,
  1995, pp. 39--43.

\bibitem{wang2021robot}
J.~Wang, J.~Chen, Y.~Sun, X.~Ma, D.~Wang, J.~Sun, and P.~Cheng, ``Robot:
  Robustness-oriented testing for deep learning systems,'' in \emph{2021
  IEEE/ACM 43rd International Conference on Software Engineering (ICSE)}.\hskip
  1em plus 0.5em minus 0.4em\relax IEEE, 2021, pp. 300--311.

\bibitem{wang2022bet}
J.~Wang, H.~Qiu, Y.~Rong, H.~Ye, Q.~Li, Z.~Li, and C.~Zhang, ``Bet: black-box
  efficient testing for convolutional neural networks,'' in \emph{Proceedings
  of the 31st ACM SIGSOFT International Symposium on Software Testing and
  Analysis}, 2022, pp. 164--175.

\bibitem{tian2018deeptest}
Y.~Tian, K.~Pei, S.~Jana, and B.~Ray, ``Deeptest: Automated testing of
  deep-neural-network-driven autonomous cars,'' in \emph{Proceedings of the
  40th international conference on software engineering}, 2018, pp. 303--314.

\bibitem{odena2019tensorfuzz}
A.~Odena, C.~Olsson, D.~Andersen, and I.~Goodfellow, ``Tensorfuzz: Debugging
  neural networks with coverage-guided fuzzing,'' in \emph{International
  Conference on Machine Learning}.\hskip 1em plus 0.5em minus 0.4em\relax PMLR,
  2019, pp. 4901--4911.

\bibitem{yu2022white}
J.~Yu, S.~Duan, and X.~Ye, ``A white-box testing for deep neural networks based
  on neuron coverage,'' \emph{IEEE Transactions on Neural Networks and Learning
  Systems}, 2022.

\bibitem{zhang2021cagfuzz}
P.~Zhang, B.~Ren, H.~Dong, and Q.~Dai, ``Cagfuzz: coverage-guided adversarial
  generative fuzzing testing for image-based deep learning systems,''
  \emph{IEEE Transactions on Software Engineering}, 2021.

\bibitem{ma2018deepgauge}
L.~Ma, F.~Juefei-Xu, F.~Zhang, J.~Sun, M.~Xue, B.~Li, C.~Chen, T.~Su, L.~Li,
  Y.~Liu \emph{et~al.}, ``Deepgauge: Multi-granularity testing criteria for
  deep learning systems,'' in \emph{Proceedings of the 33rd ACM/IEEE
  International Conference on Automated Software Engineering}, 2018, pp.
  120--131.

\bibitem{xie2022npc}
X.~Xie, T.~Li, J.~Wang, L.~Ma, Q.~Guo, F.~Juefei-Xu, and Y.~Liu, ``Npc: N euron
  p ath c overage via characterizing decision logic of deep neural networks,''
  \emph{ACM Transactions on Software Engineering and Methodology (TOSEM)},
  vol.~31, no.~3, pp. 1--27, 2022.

\bibitem{yan2020correlations}
S.~Yan, G.~Tao, X.~Liu, J.~Zhai, S.~Ma, L.~Xu, and X.~Zhang, ``Correlations
  between deep neural network model coverage criteria and model quality,'' in
  \emph{Proceedings of the 28th ACM Joint Meeting on European Software
  Engineering Conference and Symposium on the Foundations of Software
  Engineering}, 2020, pp. 775--787.

\bibitem{pavlitskaya2022neuron}
S.~Pavlitskaya, {\c{S}}.~Y{\i}km{\i}{\c{s}}, and J.~M. Z{\"o}llner, ``Is neuron
  coverage needed to make person detection more robust?'' in \emph{Proceedings
  of the IEEE/CVF Conference on Computer Vision and Pattern Recognition}, 2022,
  pp. 2889--2897.

\bibitem{zhang2020interpretingandboosting}
H.~Zhang, S.~Li, Y.~Ma, M.~Li, Y.~Xie, and Q.~Zhang, ``Interpreting and
  boosting dropout from a game-theoretic view,'' \emph{arXiv preprint
  arXiv:2009.11729}, 2020.

\bibitem{ghorbani2020neuron}
A.~Ghorbani and J.~Y. Zou, ``Neuron shapley: Discovering the responsible
  neurons,'' \emph{Advances in Neural Information Processing Systems}, vol.~33,
  pp. 5922--5932, 2020.

\bibitem{ren2021unified}
J.~Ren, D.~Zhang, Y.~Wang, L.~Chen, Z.~Zhou, Y.~Chen, X.~Cheng, X.~Wang,
  M.~Zhou, J.~Shi \emph{et~al.}, ``A unified game-theoretic interpretation of
  adversarial robustness,'' \emph{arXiv preprint arXiv:2103.07364}, 2021.

\bibitem{zhang2021building}
D.~Zhang, H.~Zhang, H.~Zhou, X.~Bao, D.~Huo, R.~Chen, X.~Cheng, M.~Wu, and
  Q.~Zhang, ``Building interpretable interaction trees for deep nlp models,''
  in \emph{Proceedings of the AAAI Conference on Artificial Intelligence},
  vol.~35, no.~16, 2021, pp. 14\,328--14\,337.

\bibitem{li2021instance}
J.~Li, K.~Kuang, L.~Li, L.~Chen, S.~Zhang, J.~Shao, and J.~Xiao,
  ``Instance-wise or class-wise? a tale of neighbor shapley for concept-based
  explanation,'' in \emph{Proceedings of the 29th ACM International Conference
  on Multimedia}, 2021, pp. 3664--3672.

\bibitem{lu2022interpretable}
W.~Lu, W.~Xu, and Z.~Sheng, ``An interpretable image tampering detection
  approach based on cooperative game,'' \emph{IEEE Transactions on Circuits and
  Systems for Video Technology}, 2022.

\bibitem{DorigoBS06}
M.~Dorigo, M.~Birattari, and T.~St{\"{u}}tzle, ``Ant colony optimization,''
  \emph{{IEEE} Comput. Intell. Mag.}, vol.~1, no.~4, pp. 28--39, 2006.

\bibitem{BurkeV97}
E.~K. Burke and D.~B. Varley, ``A genetic algorithms tutorial tool for
  numerical function optimisation,'' in \emph{Proceedings of the 2nd Annual
  Conference on Integrating Technology into Computer Science Education, ITiCSE
  1997, Uppsala, Sweden, 1-5 June, 1997}, L.~N. Cassel, M.~Daniels, J.~E.
  Miller, and G.~Davies, Eds.\hskip 1em plus 0.5em minus 0.4em\relax {ACM},
  1997, pp. 27--30.

\bibitem{szegedy2013intriguing}
C.~Szegedy, W.~Zaremba, I.~Sutskever, J.~Bruna, D.~Erhan, I.~Goodfellow, and
  R.~Fergus, ``Intriguing properties of neural networks,'' \emph{arXiv preprint
  arXiv:1312.6199}, pp. 1--10, 2013.

\bibitem{gehr2018ai2}
T.~Gehr, M.~Mirman, D.~Drachsler-Cohen, P.~Tsankov, S.~Chaudhuri, and
  M.~Vechev, ``Ai2: Safety and robustness certification of neural networks with
  abstract interpretation,'' in \emph{2018 IEEE Symposium on Security and
  Privacy (SP)}.\hskip 1em plus 0.5em minus 0.4em\relax IEEE, 2018, pp. 3--18.

\bibitem{weng2018evaluating}
T.-W. Weng, H.~Zhang, P.-Y. Chen, J.~Yi, D.~Su, Y.~Gao, C.-J. Hsieh, and
  L.~Daniel, ``Evaluating the robustness of neural networks: An extreme value
  theory approach,'' \emph{arXiv preprint arXiv:1801.10578}, pp. 1--18, 2018.

\bibitem{simonyan2014very}
K.~Simonyan and A.~Zisserman, ``\BIBforeignlanguage{English}{Very deep
  convolutional networks for large-scale image recognition},'' in
  \emph{\BIBforeignlanguage{English}{3rd International Conference on Learning
  Representations, {ICLR} 2015, San Diego, CA, USA, May 7-9, 2015}}, 2015, pp.
  1--14.

\bibitem{krizhevsky2009learning}
A.~Krizhevsky, G.~Hinton \emph{et~al.}, ``Learning multiple layers of features
  from tiny images,'' pp. 1--60, 2009.

\bibitem{russakovsky2015imagenet}
O.~Russakovsky, J.~Deng, H.~Su, J.~Krause, S.~Satheesh, S.~Ma, Z.~Huang,
  A.~Karpathy, A.~Khosla, M.~Bernstein \emph{et~al.}, ``Imagenet large scale
  visual recognition challenge,'' \emph{International journal of computer
  vision}, vol. 115, no.~3, pp. 211--252, 2015.

\bibitem{yamagishi2019cstr}
J.~Yamagishi, C.~Veaux, K.~MacDonald \emph{et~al.}, ``Cstr vctk corpus: English
  multi-speaker corpus for cstr voice cloning toolkit (version 0.92),'' pp.
  1--15, 2019.

\bibitem{he2016deep}
K.~He, X.~Zhang, S.~Ren, and J.~Sun, ``Deep residual learning for image
  recognition,'' in \emph{Proceedings of the IEEE conference on computer vision
  and pattern recognition}, 2016, pp. 770--778.

\bibitem{howard2017mobilenets}
A.~G. Howard, M.~Zhu, B.~Chen, D.~Kalenichenko, W.~Wang, T.~Weyand,
  M.~Andreetto, and H.~Adam, ``Mobilenets: Efficient convolutional neural
  networks for mobile vision applications,'' \emph{arXiv preprint
  arXiv:1704.04861}, pp. 1--9, 2017.

\bibitem{li2017deep}
C.~Li, X.~Ma, B.~Jiang, X.~Li, X.~Zhang, X.~Liu, Y.~Cao, A.~Kannan, and Z.~Zhu,
  ``Deep speaker: an end-to-end neural speaker embedding system,'' \emph{arXiv
  preprint arXiv:1705.02304}, pp. 1--8, 2017.

\bibitem{Madry2018Towards}
A.~Madry, A.~Makelov, L.~Schmidt, D.~Tsipras, and A.~Vladu, ``Towards deep
  learning models resistant to adversarial attacks,'' in \emph{6th
  International Conference on Learning Representations, {ICLR} 2018, Vancouver,
  BC, Canada, April 30-May 3, 2018}.\hskip 1em plus 0.5em minus 0.4em\relax
  OpenReview.net, 2018, pp. 1--28.

\bibitem{athalye2018obfuscated}
A.~Athalye, N.~Carlini, and D.~Wagner, ``Obfuscated gradients give a false
  sense of security: Circumventing defenses to adversarial examples,'' in
  \emph{International conference on machine learning}.\hskip 1em plus 0.5em
  minus 0.4em\relax PMLR, 2018, pp. 274--283.

\bibitem{GoodfellowSS14}
I.~J. Goodfellow, J.~Shlens, and C.~Szegedy, ``Explaining and harnessing
  adversarial examples,'' in \emph{3rd International Conference on Learning
  Representations, {ICLR} 2015, San Diego, CA, USA, May 7-9, 2015, Conference
  Track Proceedings}, Y.~Bengio and Y.~LeCun, Eds., 2015, pp. 1--11.

\bibitem{moosavi2016deepfool}
S.-M. Moosavi-Dezfooli, A.~Fawzi, and P.~Frossard, ``Deepfool: a simple and
  accurate method to fool deep neural networks,'' in \emph{Proceedings of the
  IEEE conference on computer vision and pattern recognition}, 2016, pp.
  2574--2582.

\bibitem{papernot2016limitations}
N.~Papernot, P.~McDaniel, S.~Jha, M.~Fredrikson, Z.~B. Celik, and A.~Swami,
  ``The limitations of deep learning in adversarial settings,'' in \emph{2016
  IEEE European symposium on security and privacy (EuroS\&P)}.\hskip 1em plus
  0.5em minus 0.4em\relax IEEE, 2016, pp. 372--387.

\end{thebibliography}

\end{document}